\documentclass[journal]{IEEEtran} 
\usepackage{algorithmic}
\usepackage{algorithm}
\usepackage{amsfonts}
\usepackage{amsmath}
\usepackage{amssymb}
\usepackage{balance}
\usepackage{cite}
\usepackage{graphicx}
\usepackage{lmodern}
\usepackage{mathtools}
\usepackage{multirow}
\usepackage{subfig}
\usepackage{epstopdf}
\usepackage{url}
\usepackage{color}

\DeclareMathOperator*{\argmin}{arg\,min}

\title{
Handling Imbalanced Classification Problems with Support Vector Machines via 
Evolutionary Bilevel Optimization}
\author{
\IEEEauthorblockN{Alejandro~Rosales-P\'erez, Salvador~Garc\'ia, Francisco~Herrera,~\IEEEmembership{Senior Member,~IEEE}}
\thanks{This work was partially supported by Project 
PID2020-119478GB-I00 funded by MICINN/AEI/10.13039/501100011033 and by Project 
A-TIC-434-UGR20 funded by FEDER/Junta de Andalucía-Consejería de Transformación 
Económica, Industria, Conocimiento y Universidades. The authors 
acknowledge the support from ``Laboratorio de 
Superc{\'o}mputo del Baj{\'i}o'' through project 300832 from CONACyT.}
\thanks{A. Rosales-P\'erez (alejandro.rosales@cimat.mx) is with the Centro de 
Investigaci{\'o}n en Matem{\'a}ticas (CIMAT), Monterrey, N.L., Mexico.}
\thanks{S. Garc\'ia (salvagl@decsai.ugr.es) and F. Herrera 
(herrera@decsai.ugr.es) are with the Department of Computer Science and Artificial 
Intelligence, Andalusian Research 
Institute in Data Science and Computational Intelligence (DaSCI), University of 
Granada, 18071, Granada, Spain.}
}

\begin{document}
\maketitle

\begin{abstract}
Support vector machines are popular learning algorithms to deal with binary 
classification problems. They traditionally assume equal misclassification costs 
for each class; however, real-world problems may have an uneven class distribution. 
This paper introduces EBCS-SVM: Evolutionary Bilevel Cost-sensitive Support Vector 
Machines. EBCS-SVM handles imbalanced classification problems by simultaneously 
learning the support vectors and optimizing the SVM hyper-parameters, which 
comprise the kernel parameter and misclassification costs. The resulting 
optimization problem is a bilevel problem, where the lower-level determines the 
support vectors and the upper-level the hyper-parameters. This optimization problem 
is solved using an evolutionary algorithm at the upper-level and Sequential 
Minimal 
Optimization at the lower-level. These two methods work in a nested fashion, i.e., 
the optimal support vectors help guide the search of the hyper-parameters, and the 
lower-level is initialized based on previous successful solutions. The proposed 
method is assessed using 70 datasets of imbalanced classification and compared with 
several state-of-the-art methods. The experimental results, supported by a Bayesian 
test, provided evidence of the effectiveness of EBCS-SVM when working with highly 
imbalanced datasets.
\end{abstract}


\section{Introduction}

\IEEEPARstart{S}{upport} vector machines (SVMs)~\cite{Cortes1995} are among the 
most popular 
supervised learning algorithms, with strong theoretical foundations and high 
effectiveness in real-world problems. The idea behind SVMs is to find the 
hyper-plane that maximizes the separation margin between two categories. In their 
canonical form, SVMs assume an equal cost for each class. This assumption works 
well when the number of instances for each class is roughly similar. However, 
real-world problems seldom have a balanced class distribution.

The imbalanced classification problem refers to an uneven class 
distribution~\cite{Fernandez2018, He2009, Krawczyk2016}, i.e., there is an 
over-represented class, known as the majority class, and an under-represented 
class, known as the minority class. Imbalanced classification problems further
have the characteristic that the minority class is the one of interest. Therefore, 
accurately recognizing the minority class becomes crucial in several applications, 
such as medical diagnosis, fraud detection, and face recognition.

There are two main approaches to deal with imbalanced classification problems with 
SVMs: data-level and algorithm-level methods~\cite{Zhu2020}. The first approach 
aims to balance the dataset through oversampling the minority 
class~\cite{He2009, Chawla2002, Nguyen2011, He2008} or undersampling the majority 
class~\cite{Hart1968, He2009, Kang2017, Xia2021}. Then, SVM learns from the edited 
dataset~\cite{Nalepa2019}. Although data-level methods are flexible, they ignore 
the particularities of learning algorithms. Conversely, algorithm-level methods  
modify the learning algorithm to be robust to uneven class distributions. For 
SVMs, common modifications comprise hyper-plane shifting~\cite{Tasadduq2006, 
Datta2015}, kernel adaptation~\cite{Mathew2018}, and 
cost-sensitive~\cite{Datta2019}. Algorithm-level methods often offer better 
performance than data-level ones~\cite{Datta2019}; however, they need to define a 
set of hyper-parameters, such as the extent of shifting compensation or the correct 
costs\footnote{The misclassification costs are the weights applied to errors 
incurred by classifying positive or negative samples.} for each class. 
Therefore, methods for imbalanced classification must 
not only learn 
when class distributions are unequal, but their hyper-parameters must also be 
tuned to get peak 
performance.

Bilevel optimization (BLO) arises as an alternative for hyper-parameter 
optimization. BLO differs from traditional optimization in that the optimization 
problem has as part of its constraints a second optimization problem. In our 
context, the principal optimization or upper-level problem is the hyper-parameter 
optimization, and the second or lower-level problem is the learning of support 
vectors. These two problems interplay, i.e., the definition of the 
hyper-parameters influences the optimal set of support vectors, and this 
set of support vectors defines the 
model to predict unseen cases. However, BLO problems are computationally 
challenging because of their non-convexity and non-linearity~\cite{Sinha2017, 
Sinha2018}.

Evolutionary algorithms (EAs) are powerful search tools capable of solving complex 
optimization problems, such as BLO problems. Recently, the interest in using EAs 
to 
address machine learning problems is growing fastly~\cite{Chapelle2002, Dantas2016, 
Li2010, evoML3, Mierswa2006, RosalesPerez2018, evoML1, evoML2, RosalesPerez2015, 
RosalesPerez2017, Stanley2019}. 
For imbalanced learning, EAs have been used for data sampling~\cite{Le2021,Li2021} 
and cost-sensitive learning~\cite{Zhang2019}. 
Although recent studies address the problem of determining the optimal 
misclassification costs~\cite{Wang2021, Zhang2019}, they have paid little attention 
to considering the hyper-parameters of the learning algorithm, along with 
exploiting the hierarchical nature of parameter and hyper-parameter learning to 
guide search. Furthermore, taking advantage of the properties of the learning 
algorithm to estimate the classification performance efficiently in imbalanced 
problems is almost unexplored.

In the light of the above mentioned, this paper introduces EBCS-SVM: Evolutionary 
Bilevel Cost-Sensitive Support Vector 
Machines. EBCS-SVM combines an EA and the Sequential Minimal Optimization 
(SMO) algorithm in a nested manner. The EA optimizes the cost of hyper-parameters, 
which are the costs of each class and the kernel parameters, and the SMO learns the 
optimal support vectors. These two optimizers interact such that information from 
one level is used by the other to improve search and convergence capabilities. We 
summarize the main contributions of this paper as follows:

\begin{itemize}
\item {We propose EBCS-SVM, which allows learning SVMs in imbalanced 
classification problems and automatically sets the misclassification costs and 
kernel parameter.} 
\item EBCS-SVM uses the information from the lower-level to guide the 
search to the upper-level and takes advantage of the previous successful 
hyper-parameters to initialize the set of support vectors.
\item The bilevel formulation that jointly learns parameters and 
hyper-parameters.
\item The definition of the upper-level objective function that allows 
estimating the performance of the SVM without performing cross-validation. 
\end{itemize} 

The performance of EBCS-SVM was assessed using a suite of 70 benchmark datasets 
of imbalanced classification and compared with the state-of-the-art methods. 
The experimental evaluation revealed an outstanding efficacy of EBCS-SVM when 
faced with problems with a high disproportion of classes. 
The hierarchical Bayesian test supported the main findings.

The rest of this paper is organized as follows. 
Section~\ref{sec:related_work} introduces the related works 
on imbalanced 
classification problems, hyper-parameter optimization, and bilevel optimization. 
Section~\ref{sec:bilevel_svm} describes the general bilevel formulation for 
cost-sensitive SVM. Next, Section~\ref{sec:ebcs2vm} describes the proposed 
EBCS-SVM. Section~\ref{sec:experimental_settings} details the datasets, 
reference methods, and performance measures, while 
Section~\ref{sec:experiments_results} presents the experimental results. 
Finally, Section~\ref{sec:conclusions} discusses the main conclusions.

\section{Related Work} \label{sec:related_work}

This section presents the preliminaries. 
Section~\ref{sec:imbalance_classification} describes the main approaches for 
imbalanced classification. Then, Section~\ref{sec:hyper_parameters} describes the 
hyper-parameter optimization problem, and 
Section~\ref{sec:evolutionary_bilevel} presents the main concepts related to 
bilevel optimization.

\subsection{Methods for Imbalance Classification} 
\label{sec:imbalance_classification} 

Most learning algorithms may face difficulties when dealing with imbalanced 
classification problems, as they can favor the majority class, leading to an 
ineffective classification model. Two main approaches to dealing with imbalanced 
datasets are sampling strategies and algorithm 
adaptation~\cite{Fernandez2018,He2009, Krawczyk2016}. The former works with 
training data by modifying its class distribution, while the latter adjusts the 
training algorithm or inference process to consider the imbalance. We describe 
these two approaches below.

\subsubsection{Data-Level Preprocessing Methods}

This approach aims to reduce the effect of class imbalance by adding or 
removing samples from training data to balance the class 
distribution~\cite{Fernandez2018SMOTE}. There are two primary sampling 
strategies:

\begin{itemize}
\item {\bf Oversampling} methods attempt to balance the dataset by replicating or 
creating samples from the minority class. 
Random oversampling (ROS)~\cite{He2009} is the simplest method for data 
balancing that replicates samples from the minority class. SMOTE~\cite{Chawla2002} 
is a popular method for generating artificial samples through a linear 
interpolation of samples from the minority class. Variants of SMOTE include 
SVMSMOTE~\cite{Nguyen2011} and ADASYN~\cite{He2008}.

The major criticism of oversampling methods is that the synthetic samples can 
cause overfitting of the classification model.

\item {\bf Undersampling} methods balance the dataset by removing instances from 
the majority class. Random undersampling (RUS)~\cite{He2009} is an uninformed 
method that removes instances from the majority class at random. The condensed 
nearest neighbor (CNN)~\cite{Hart1968} is an informed method that eliminates 
examples distant from the boundary decision.

The major criticism of undersampling methods is that they can discard 
meaningful instances and lead to loss of information.
\end{itemize}

It is unclear whether oversampling is better than undersampling or vice 
versa~\cite{Estabrooks2004, LOPEZ20126585}. Both methods are effective in 
handling imbalanced classification problems.

\subsubsection{Algorithm-Level Methods}

This approach aims to adapt the way a particular classifier learns in such a 
manner that it can deal with imbalanced problems. For SVMs, there are three main 
approaches of adaptation:

\begin{itemize}
\item {\bf Cost-sensitive} methods consider different costs to each class during 
learning, such that minority class errors have a higher penalization than those of 
the majority class. SVMs can work in a cost-sensitive framework by using different 
regularization parameters for positive and negative samples~\cite{Veropoulos1999}. 
Also, an instance can be weighted based on the density of its 
neighborhood~\cite{Hazarika2021}. However, the proper setting of the costs is 
unknown.

\item {\bf Kernel adaptation} methods adapt the kernel function or kernel 
matrix to reduce the bias towards the majority class. For example, 
WK-SMOTE~\cite{Mathew2018} expands the kernel matrix by incorporating the dot 
products of artificial samples generated in the feature space. Then, the SVM 
training algorithm uses the modified kernel matrix to learn a model.
\item {\bf Hyper-plane shifting} methods shift the separating hyper-plane to 
enlarge the margin around minority class~\cite{Datta2015}.
\end{itemize}

The major criticism of the algorithm-level methods is that they are 
algorithm-specific and require in-depth knowledge of the classifier. However, 
these methods are more accurate than data-level methods~\cite{Fernandez2018}.

\subsection{Hyper-Parameter Optimization} \label{sec:hyper_parameters}

Hyper-parameter optimization refers to the problem of automatically setting the 
hyper-parameter configuration of a learning algorithm to optimize the performance. 
Hyper-parameter optimization is a complicated problem with several challenges. The 
challenges include computationally expensive evaluations of the objective function, 
a complex and non-convex search space, hyper-parameters that cannot be 
differentiable, and a finite amount of data that may limit the estimation of the 
generalization performance~\cite{Feurer2019}. Formally, the hyper-parameter 
optimization problem can be stated as follows~\cite{Feurer2019}:

\begin{equation}
\mathbf{\theta}^\ast = \text{arg }\min_{\theta \in \Theta} 
\mathbb{E}_{D_{train}, D_{val}} \mathcal{L}\left(
\mathcal{A}_\theta, D_{train}, D_{val} \right)
\label{eq:hpo}
\end{equation}
\noindent where $\mathcal{L}\left(\mathcal{A}_\theta, D_{train}, D_{val} 
\right)$ is a loss function that measures the loss of the model learned by 
algorithm $\mathcal{A}$ with hyper-parameters $\theta$ that is trained with 
$D_{train}$ and is validated with $D_{val}$.

Global optimization techniques are commonly adopted to face the non-convex 
nature of the hyper-parameter optimization problem. For SVMs, the works on 
hyper-parameter optimization can be categorized into:
\begin{itemize}
\item {\bf Model-free optimization} includes classical techniques such as Grid 
Search~\cite{Bergstra2012}, Random Search~\cite{Bergstra2012}, Evolutionary 
Algorithms~\cite{Friedrich2005, Rosales2014b}, and Particle Swarm 
Optimization~\cite{Li2010}. 
\item {\bf Model-based optimization} includes Bayesian 
optimization~\cite{Feurer2015, Thornton2013} and surrogate-assisted 
optimization~\cite{RosalesPerez2015}.
\end{itemize}

Works on SVMs hyper-parameter optimization have also considered the 
optimization of the model pipeline (data preprocessing + learning 
algorithm)~\cite{AlZoubi2021, Rosales2014b}, the training set selection 
problem~\cite{Acampora2018,Cheng2020,RosalesPerez2017}, or more recently a 
combination of feature and training set selection together with hyper-parameter 
optimization~\cite{Dudzik2021}. However, these studies neglected the imbalanced 
classification problem.

\subsection{Evolutionary Bilevel Optimization} \label{sec:evolutionary_bilevel}

Bilevel optimization is a hierarchical optimization problem with two 
levels: the upper-level, also known as the leader, and the lower-level, also 
called the follower. Formally, a bilevel optimization problem is stated as 
follows~\cite{Sinha2017, Sinha2018}:

\begin{equation}
\begin{aligned}
 \min & f_u\left(\mathbf{v}_u,\mathbf{v}_l\right)\\
 \text{s.t. } & \mathbf{v}_l \in \left\lbrace \argmin 
f_l\left(\mathbf{v}_u,\mathbf{v}_l: g_l\left(\mathbf{v}_u,\mathbf{v}_l\right) 
\leq 0\right) \right\rbrace\\ 
 & g_u\left(\mathbf{v}_u,\mathbf{v}_l\right) \leq 0
\end{aligned}
\end{equation}

\noindent where $\mathbf{v}_u$ and $\mathbf{v}_l$ are the upper-level and the
lower-level variables, respectively, $f_u$ and $f_l$ represent the objective 
functions for the upper-level and lower-level, and $g_u$ and $g_l$ are the set of 
constraints for upper-level and lower-level, respectively. 

The lower-level is a constraint to the upper-level optimization problem. 
Therefore, the lower-level solution partially determines the upper-level solution. 
The nested structure leads to several difficulties, such as non-linearity, 
non-convexity, and disconnectedness. These difficulties can be present even for 
the simplest bilevel problems~\cite{Sinha2017, Sinha2018,Sinha2021}. 

EAs have shown success when dealing with complex optimization problems. Thus, 
EAs emerge as an alternative to deal with bilevel optimization problems. Most 
of the current EAs proposed to handle these problems are nested in nature. 
These approaches have two optimization algorithms, where one algorithm runs 
within the other. Overviews of evolutionary bilevel algorithms can be found 
in~\cite{Sinha2017, Sinha2018}.

Supervised learning can be treated as a bilevel problem, in which the upper-level 
optimizes the hyperparameters that minimize the expected generalization error, and 
the lower-level learns the parameters~\cite{Dhebar2020}. Next, we explain the 
bilevel formulation for learning parameters and 
hyper-parameters for an SVM with cost-sensitive.

\section{Bilevel Cost-Sensitive Support Vector Machine: Optimization Problem} 
\label{sec:bilevel_svm}

The bilevel formulation breaks the problem down into two levels: the upper-level 
concerned with the hyper-parameters configuration and the lower-level with the SVM 
training. In this section, we explain the optimization objectives at each level. 
First, Section~\ref{sec:lower} defines the lower-level that finds the optimal 
separating hyper-plane when training the SVM. Next, Section~\ref{sec:upper} 
explains the objective function for the upper-level, which optimizes the 
classification performance considering the uneven class distributions for a given 
hyper-parameter 
configuration.

\subsection{Lower-Level -- Optimizing Parameters} \label{sec:lower}

The lower-level problem focuses on finding the support vectors that define the 
hyper-plane for an imbalanced classification problem. A cost-sensitive SVM 
penalizes the errors differently for positive and negative classes. 
This is  formulated as follows~\cite{Veropoulos1999}:

\begin{equation}
\begin{aligned}
  \min & \dfrac{1}{2} \parallel \mathbf{w} \parallel^2 + C^{+}\sum_{i:y=1} 
\xi_i 
+ C^{-}\sum_{i:y = -1} \xi_i\\ 
  \text{subject to } & y_i\left(\langle \mathbf{w}, \mathbf{x}_i \rangle + b 
\right) \geq 1 - \xi_i\\\ 
  & \xi_i \geq 0
\end{aligned}
\end{equation}

\noindent where $\langle \cdot,\cdot\rangle$ represents the dot product, 
$\mathbf{w}$ is the separating hyper-plane, and $C^+$ and $C^-$ are the costs 
for the positive and negative samples, respectively.  

The dual problem obtained through Lagrange multipliers is as follows:

\begin{equation}
 \begin{aligned}
  \max & \sum_{i = 1}^n \alpha_i - \dfrac{1}{2} \sum_{i,j}^n 
\alpha_i\alpha_jy_iy_j\left\langle \mathbf{x}_i,\mathbf{x}_j\right\rangle\\  
\text{subject to } & \sum_{i=1}^n y_i\alpha_i = 0\\ 
  & 0 \leq \alpha_i \leq C^+, i:y=+1\\\
  & 0 \leq \alpha_i \leq C^-, i:y=-1
\end{aligned}
\label{eq:lowerLevel}
\end{equation}

For learning nonlinear functions, the dot product  $\langle 
\mathbf{x}_i,\mathbf{x}_j \rangle$ is replaced by a kernel function $K\left( 
\mathbf{x}_i,\mathbf{x}_j \right)$. {The Radial basis function (RBF) is a 
kernel function that is highly effective and theoretically 
supported~\cite{Bagnall201,Minh2010}. The RBF kernel is defined as:}

\begin{equation}
 K\left(\mathbf{x}_i,\mathbf{x}_j\right) = 
\mathrm{e}^{-\gamma\|\mathbf{x}_i-\mathbf{x}_j\|^2}
\end{equation}

\noindent where $\gamma$ is an adjustable parameter given by the upper-level.

Section~\ref{sec:upper} explains the upper-level optimization problem that 
optimizes the hyper-parameters $C^+$, $C^-$, and the $\gamma$ value for the RBF 
kernel.

\subsection{Upper-Level -- Optimizing Hyper-Parameters} \label{sec:upper}

The upper-level optimizes the hyper-parameters used in the lower-level to learn 
the support vectors. Let $\lambda$ be a vector that encodes the 
hyper-parameters $C^+$, $C^-$, and $\gamma$. The goal of the upper-level is to find 
the set of hyper-parameters that gets the minimum generalization error on 
imbalanced classification problems, which is estimated using the balanced error 
rate (BER) score. The BER is defined as:

\begin{equation}
BER \left( \lambda, \alpha^*, b \right)  = \dfrac{1}{2}\left( 
\dfrac{FN}{TP+FN} + \dfrac{FP}{FP+TN} \right)
\label{eq:BER}
\end{equation}

\noindent where $TP$ and $TN$ are, respectively, the number of positive and 
negative samples correctly classified; and $FN$ and $FP$ are, respectively, the 
number of positive and negative samples incorrectly classified.

Computing BER using the training set can lead to overfitting. $K$-fold 
cross-validation is commonly used to assess the expected performance and to reduce 
the risk of overfitting. However, this procedure can become computationally 
inefficient, as it implies solving the lower-level problem $K$ times for each 
configuration of hyper-parameters $\lambda$. We face that disadvantage by 
approximating the bound on the leave-one-out cross-validation for the 
upper-level. Based on the lower-level solution $\alpha^*$, the predicted value 
for the $j^{th}$ training sample is given by:

\begin{equation}
 f\left( \mathbf{x}_j \right) = \sum_{i=1}^n \alpha_i^* y_i K\left( 
\mathbf{x}_j, 
\mathbf{x}_i \right) + b
\label{eq:svm_lower_level}
\end{equation}

\noindent where $b$ is the bias term and is set to satisfy the Karush-Kuhn-Tucker 
condition.

Eliminating a non-support vector from the training set does not affect the model; 
therefore, we focus on support vectors, as they can contribute to the error. 
Assuming that the set of support vectors remains the same during the 
leave-one-out procedure, we can approximate the output when removing the 
$j^{th}$ support vector from the training set as:

\begin{equation}
 \hat{f}\left( \mathbf{x}_j \right) = \sum_{i=1}^n \alpha^*_i y_i K\left( 
\mathbf{x}_j, \mathbf{x}_i \right) + b - \alpha^*_j y_j K\left( \mathbf{x}_j, 
\mathbf{x}_j \right)
\label{eq:boundApproximation_original}
\end{equation}

In the case of the RBF kernel, the term $K\left( \mathbf{x}_j, \mathbf{x}_j 
\right)$ equals one. Simplifying 
(\ref{eq:boundApproximation_original}) and multiplying it by $y_j$, an 
instance $ \mathbf{x}_j$ is incorrectly classified if:

\begin{equation}
 y_j\left(\sum_{i=1}^n \alpha^*_i y_i K\left( 
\mathbf{x}_i, \mathbf{x}_j \right) + b - \alpha^*_j y_j \right) < 0
\label{eq:boundApproximation}
\end{equation}

After determining the incorrectly classified instances with 
(\ref{eq:boundApproximation}), the BER of each individual at the upper-level is 
determined using (\ref{eq:BER}).

After solving the bilevel optimization problem, the optimal support vectors are 
used to classify a new instance using (\ref{eq:svm_lower_level}).

\section{EBCS-SVM: Evolutionary Bilevel Cost-Sensitive Support Vector 
Machines} \label{sec:ebcs2vm}

\begin{figure*}[!ht]
\centering
 \includegraphics{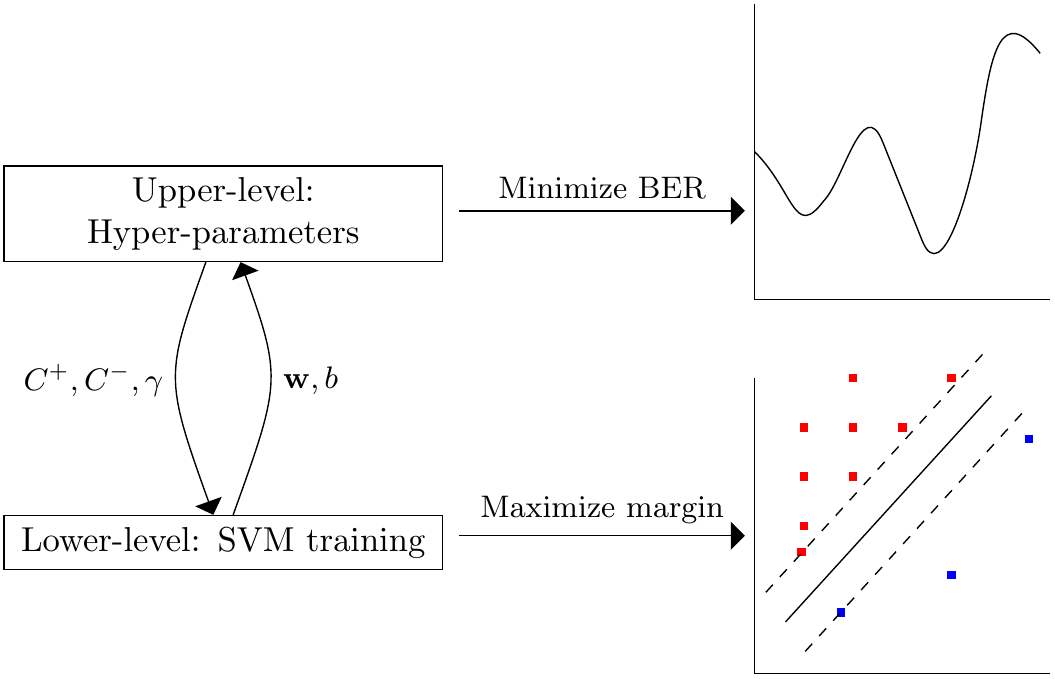}
 \caption{Bilevel scheme to learn support vectors machine in a cost-sensitive 
approach. The figure depicts the dependency between hyper-parameters and support 
vectors, optimizing the margin with different penalties. }
\label{fig:bilevelInteraction}
\end{figure*}

EBCS-SVM aims to build an optimal SVM model for handling imbalanced 
classification problems through bilevel optimization. EBCS-SVM determines the 
hyper-parameters values, i.e., the costs of each class and the kernel parameters 
that minimize BER at the upper-level, and the lower-level finds the optimal 
separating hyper-plane. Fig.~\ref{fig:bilevelInteraction} graphically depicts the 
bilevel interaction between the lower-level and upper-level in EBCS-SVM.

\begin{algorithm}
\caption{EBCS-SVM}
\label{alg:ebcs2vm}
\begin{algorithmic}[1]
\begin{small}
\REQUIRE $\mathcal{X}$, the set of samples,\\ $y$, the set of 
classes 
labels,\\$u_s$, upper population size,\\ $m_e$, 
maximum number of evaluations at the upper-level, \\ $\tau$, tolerance 
threshold for lower-level, \\ $m$, maximum number of iterations for 
lower-level. 
\ENSURE The set of support vectors
\STATE Generate randomly an initial population of hyper-parameters, 
$\mathcal{P}_u$, with $u_s$ 
individuals 
\FOR {each individual $p_u$ in $\mathcal{P}_u$}
\STATE Use SMO with the hyper-parameters defined in $p_u$ to find the optimal support vectors.
\STATE Compute the BER for the upper-level individual using the optimal support 
vectors.
\ENDFOR
\WHILE {a stopping criterion is not met}
\STATE Apply evolutionary operator to produce an upper-level offspring 
population, $\mathcal{O}_u$.
\FOR {each individual $o_u$ in $\mathcal{O}_u$}
\STATE Find the nearest neighbors of $o_u$ in $\mathcal{P}_u$.
\STATE Warm starting the set of support vectors of the lower-level based on the 
nearest neighbors.
\STATE Use SMO with the hyper-parameters defined in $o_u$ to find the optimal support vectors.
\STATE Compute the BER for the upper-level offspring using the optimal support 
vectors.
\ENDFOR
\ENDWHILE
\STATE Choose the solution with the lowest BER as the optimal SVM for handling 
imbalanced datasets. If there are more than one solution with similar BER to the 
lowest one, choose the one with the lower number of support vectors.
\end{small}
\end{algorithmic}
\end{algorithm}

Algorithm~\ref{alg:ebcs2vm} describes EBCS-SVM, and it works as follows:

\begin{enumerate}
\item In line 1, a population for the upper-level is randomly created with $u_s$ 
individuals and three variables representing the costs for the positive and the 
negative class, and the $\gamma$ value for RBF kernel, based on the bounds given in 
Section~\ref{sec:evo_op}.
\item In lines 2--5, each upper-level solution is evaluated. To this end, the 
following steps are carried out:
\begin{enumerate}
 \item The lower-level problem described in Section~\ref{sec:lower} is solved 
using the SMO algorithm and the given hyper-parameters.
\item The resulting SVM model is evaluated by computing the BER, as described in 
Section~\ref{sec:upper}.
\end{enumerate}
\item In lines 6--14, the evolutionary process takes place.
\begin{enumerate}
\item In line 7, a new population of hyper-parameters is created by applying 
evolutionary operators over $\mathcal{P}_u$ using the adaptation 
of DE operators proposed by SHADE~\cite{Tanabe2013}.
\item In line 9, the nearest neighbors in $\mathcal{P}_u$ are found for each 
individual in $\mathcal{O}_u$. In line 10, the set of support vectors for the 
lower-level is warm started by considering the nearest neighbors of the 
upper-level 
to determine how probable an instance is a support vector, as described in 
Section~\ref{sec:lower_level_initialization}.
\item Line 11 solves the lower-level optimization problem with the SMO 
algorithm and, line 12 computes the BER score for the given hyper-parameters.
\end{enumerate}
\end{enumerate}

\subsection{Lower-level Initialization} \label{sec:lower_level_initialization}

When solving the lower-level using the hyper-parameter of the initial 
upper-level population, all training instances are initially assumed to be a 
support vector, and the SMO solves the lower-level problem. The solutions obtained 
by the initial configuration of hyper-parameters are stored to perform a 
warm starting of the lower-level.

The warm starting takes place after the first generation of the 
upper-level and works as follows. First, for a new given hyper-parameters 
configuration, its $m$ nearest-neighbors are found among the hyper-parameters from 
the current population. After, the set of support vectors is retrieved and used 
to determine the chance of an instance becoming a support vector, based on 
the relative frequency of its 
$m$ nearest-neighbors. 
Finally, the SMO algorithm solves the lower-level based on previous 
initialization. The premise for using this initialization is that similar 
configurations of hyper-parameters lead to a similar set of support vectors.

The value of $m$ is determined during the search. For doing so, the number of 
neighbors is randomly selected between $\left[1, u_s\right]$ with uniform 
probability. Then, the probability of each value of $m$ is updated based on the 
normalized frequency of success of such value, i.e., an improvement in the BER 
score.

\subsection{Representation and Evolutionary Operators} \label{sec:evo_op}

At the upper-level, individuals encode the hyper-parameters as a real-value 
vector. Initially, $C^+$ and $C^-$ are randomly generated in the range $\left[ 
2^{-5} , 2^{10}\right]$ and $\gamma$ between $\left[ 2^{-10}, 2^{5} \right]$.
Individuals at this level are then evolved using DE~\cite{Price2005book, 
Storn1997}. However, the classical DE requires the definition of two parameters: 
the differential weights ($F$) and the crossover rate ($CR$). For this reason, we 
adopt a self-adaptative variant called SHADE~\cite{Tanabe2013}, which 
uses 
history information to adapt the values of $F$ and $CR$ during the search and a 
greedy current-to-best mutation strategy. Therefore, these parameters are learned 
during the search.

After a child solution of hyper-parameters is created and its BER value is 
computed, it competes with the current individual to determine which one is better 
and therefore survives. A child solution is better if its BER value is lower than 
the BER value of the current individual, or if the BER values of both solutions are 
equivalent, but the child has fewer support vectors. 

\section{Experimental Settings} \label{sec:experimental_settings}

In this section, we describe the configuration setup for our experimental 
study. In Section~\ref{sec:datasets}, we present the set of benchmark datasets 
considered for experimentation. Section~\ref{sec:statistical} provides the 
metrics used to assess the performance of all methods and the test to analyze 
them. Finally, Section~\ref{sec:reference_methods} details the state-of-the-art 
techniques used to compare the performance and their 
respective tuning of hyper-parameters.

\subsection{Datasets} \label{sec:datasets}

The performance of EBCS-SVM is assessed using a benchmark of 70 datasets from the 
KEEL repository~\cite{KEEL}. 
Table~\ref{tab:datasets_description} details 
the characteristics of the datasets, including the number of instances (Inst.), 
the number of features (Feat.), and the imbalance ratio (IR)\footnote{{IR is 
defined as the 
ratio between the number of samples in the majority class and the number of 
samples 
in the minority. Therefore, the higher the IR, the greater the imbalance.}}. These 
datasets are 
diverse in the IR, number of samples, and number of features. Thus, the 
performance is assessed using problems with different 
characteristics.
We divided the datasets into three groups 
based on the degree of IR: ten 
datasets with small IR, i.e., the IR is less than or equal to three; 35 datasets 
with 
medium IR, that is an IR higher than three and less than or equal to 20; and 25 
datasets with high IR, i.e., the IR is higher than 20.

\begin{table}
 \caption{Description of datasets based on the number of instances (Inst.), 
number of features (Feat.), and the imbalance ratio (IR).} 
\label{tab:datasets_description}
\ContinuedFloat
\resizebox{0.495\textwidth}{319.5pt}{
\begin{tabular}{llrrr}
\hline
ID & Dataset	&	Inst.	&	Feat.	&	IR	\\
\hline
\multicolumn{4}{c}{Small IR}\\
1	&	ecoli-0\_vs\_1	&	220	&	7	&	1.857	\\
2	&	glass0	&	214	&	9	&	2.057	\\
3	&	glass1	&	214	&	9	&	1.816	\\
4	&	haberman	&	306	&	3	&	2.778	\\
5	&	iris0	&	150	&	4	&	2.000	\\
6	&	pima	&	768	&	8	&	1.866	\\
7	&	vehicle1	&	846	&	18	&	1.029	\\
8	&	vehicle2	&	846	&	18	&	2.881	\\
9	&	vehicle3	&	846	&	18	&	1.029	\\
10	&	wisconsin	&	683	&	9	&	1.858	\\
	\multicolumn{4}{c}{Medium IR}\\
11	&	abalone9-18	&	731	&	8	&	16.405	\\
12	&	cleveland-0\_vs\_4	&	173	&	13	&	12.308	\\
13	&	dermatology-6	&	358	&	34	&	16.900	\\
14	&	ecoli-0-1-4-6\_vs\_5	&	280	&	6	&	13.000	\\
15	&	ecoli-0-1-4-7\_vs\_2-3-5-6	&	336	&	7	&	10.586	\\
16	&	ecoli-0-1-4-7\_vs\_5-6	&	332	&	6	&	12.280	\\
17	&	ecoli-0-1\_vs\_2-3-5	&	244	&	7	&	9.167	\\
18	&	ecoli-0-3-4-7\_vs\_5-6	&	257	&	7	&	9.280	\\
19	&	ecoli-0-3-4\_vs\_5	&	200	&	7	&	9.000	\\
20	&	ecoli-0-6-7\_vs\_5	&	220	&	6	&	10.000	\\
21	&	ecoli1	&	336	&	7	&	3.364	\\
22	&	ecoli2	&	336	&	7	&	5.462	\\
23	&	ecoli3	&	336	&	7	&	8.600	\\
24	&	ecoli4	&	336	&	7	&	15.800	\\
25	&	glass-0-1-2-3\_vs\_4-5-6	&	214	&	9	&	3.196	\\
26	&	glass-0-1-5\_vs\_2	&	172	&	9	&	9.118	\\
27	&	glass-0-1-6\_vs\_5	&	184	&	9	&	19.444	\\
28	&	glass-0-4\_vs\_5	&	92	&	9	&	9.222	\\
29	&	glass-0-6\_vs\_5	&	108	&	9	&	11.000	\\
30	&	glass2	&	214	&	9	&	11.588	\\
31	&	glass4	&	214	&	9	&	15.462	\\
32	&	glass6	&	214	&	9	&	6.379	\\
33	&	led7digit-0-2-4-5-6-7-8-9\_vs\_1	&	443	&	7	&	8.630	\\
34	&	new-thyroid1	&	215	&	5	&	5.143	\\
35	&	newthyroid2	&	215	&	5	&	5.143	\\
36	&	page-blocks-1-3\_vs\_4	&	472	&	10	&	15.857	\\
37	&	segment0	&	2308	&	19	&	6.015	\\
38	&	shuttle-c0-vs-c4	&	1829	&	9	&	13.870	\\
39	&	vehicle0	&	846	&	18	&	3.251	\\
40	&	vowel0	&	988	&	13	&	9.978	\\
41	&	yeast-0-3-5-9\_vs\_7-8	&	506	&	8	&	9.120	\\
42	&	yeast-0-5-6-7-9\_vs\_4	&	528	&	8	&	9.353	\\
43	&	yeast-1\_vs\_7	&	459	&	7	&	14.300	\\
44	&	yeast-2\_vs\_4	&	514	&	8	&	9.078	\\
45	&	yeast3	&	1484	&	8	&	8.104	\\
\multicolumn{4}{c}{High IR}\\
46	&	abalone-17\_vs\_7-8-9-10	&	2338	&	8	&	39.310	\\
47	&	abalone-20\_vs\_8-9-10	&	1916	&	8	&	72.692	\\
48	&	abalone-21\_vs\_8	&	581	&	8	&	40.500	\\
49	&	abalone-3\_vs\_11	&	502	&	8	&	32.467	\\
50	&	abalone19	&	4174	&	8	&	129.438	\\
51	&	car-good	&	1728	&	6	&	24.043	\\
52	&	flare-F	&	1066	&	11	&	23.791	\\
53	&	glass5	&	214	&	9	&	22.778	\\
54	&	poker-8-9\_vs\_5	&	2075	&	10	&	82.000	\\
55	&	poker-8-9\_vs\_6	&	1485	&	10	&	58.400	\\
56	&	poker-8\_vs\_6	&	1477	&	10	&	85.882	\\
57	&	poker-9\_vs\_7	&	244	&	10	&	29.500	\\
58	&	shuttle-6\_vs\_2-3	&	230	&	9	&	22.000	\\
59	&	winequality-red-3\_vs\_5	&	691	&	11	&	68.100	\\
60	&	winequality-red-4	&	1599	&	11	&	29.170	\\
61	&	winequality-red-8\_vs\_6-7	&	855	&	11	&	46.500	\\
62	&	winequality-red-8\_vs\_6	&	656	&	11	&	35.444	\\
63	&	winequality-white-3-9\_vs\_5	&	1482	&	11	&	58.280	\\
64	&	winequality-white-3\_vs\_7	&	900	&	11	&	44.000	\\
65	&	yeast-1-2-8-9\_vs\_7	&	947	&	8	&	30.567	\\
66	&	yeast-1-4-5-8\_vs\_7	&	693	&	8	&	22.100	\\
67	&	yeast-2\_vs\_8	&	482	&	8	&	23.100	\\
68	&	yeast4	&	1484	&	8	&	28.098	\\
69	&	yeast5	&	1484	&	8	&	32.727	\\
70	&	yeast6	&	1484	&	8	&	41.400	\\
\hline
\end{tabular}
}
\end{table}

Datasets were partitioned using the $10 \times 5$-fold cross-validation. In the 
$5$-fold cross-validation, the dataset is randomly split into five disjoint 
subsets. In each fold, a subset is used as the test set and the remaining as the 
training set. Then, $5$-fold cross-validation is repeated ten times, with a 
different split each time. Thus, each dataset is tested $50$ times.

\subsection{Performance Metrics and Statistical Tests}\label{sec:statistical}

We assessed the performance of the methods using a set of metrics well-suited for 
imbalanced classification problems~\cite{Luque2019}.
Let sensitivity ($sen$) and specificity ($spe$) be defined as:
\begin{equation}
\begin{aligned}
 sen &= \dfrac{TP}{TP + FN}\\
 spe &= \dfrac{TN}{TN + FP}
\end{aligned}
\end{equation}

\noindent where $TP$ and $TN$ are, respectively, the number of positive and 
negative samples correctly classified; and $FN$ and $FP$ are, respectively, the 
number of positive and negative samples incorrectly classified.

{The metrics described in~\cite{Luque2019} can be defined in terms of 
specificity and sensitivity. These metrics are listed below:}
\begin{itemize}
 \item {\bf BAR}, the balanced accuracy rate is equivalent to 1-BER and is 
defined as the average between sensitivity and specificity, {i.e.,
\begin{equation}
\text{BAR} = \dfrac{1}{2}\left(sen + spe\right)
\end{equation}}
\item {\bf BMI}, the bookmarker informedness is defined as the sum of 
sensitivity and specificity minus one, {i.e.,
\begin{equation}
\text{BMI} = sen + spe - 1
\end{equation}
}
\item {\bf GM} indicates the geometric mean between sensitivity and specificity, 
{i.e.,
\begin{equation}
\text{GM} = \sqrt{sen \cdot spe}
\end{equation}
}
\item {\bf uF1}~\cite{Luque2019} is the unbiased version of the F1 score and is 
defined as two times the sensitivity between the sum of two plus sensitivity minus 
specificity, {i.e.,
\begin{equation}
\text{uF1} = \dfrac{2 \cdot sen}{2 + sen - spe}
\end{equation}
}
\item {\bf uMCC}~\cite{Luque2019} is the unbiased version of the Matthews 
correlation coefficient, {and is defined as follows:
\begin{equation}
\text{uMCC} = \dfrac{sen \cdot spe - 1}{\sqrt{1 - \left( sen - spe\right)^2}}
\end{equation}
}
\end{itemize}

The Bayesian hypothesis tests are used to analyze EBCS-SVM regarding reference 
methods. They allow comparing the difference in the results achieved by two 
algorithms, estimating the posterior probabilities that one algorithm is better 
than the other and that both are practically equivalent. These tests are not 
affected by the number of datasets. Moreover, they can provide more information 
than the null hypothesis significance test, even when the latter does not reject 
the null hypothesis~\cite{Benavoli2017, Corani2017}. We used the hierarchical 
Bayesian test to analyze the results, as it considers both the mean and the 
variance through the cross-validation partitions for each dataset. In the analysis, 
we considered that two methods are equivalent if the difference is below 0.01.

\subsection{Reference Methods} \label{sec:reference_methods}

{We compared EBCS-SVM with several state-of-the-art techniques for 
imbalanced 
classification. To this end, we considered methods for both data-level 
preprocessing and algorithm-level. Specifically, the comparative study 
considers the following methods.
\begin{itemize}
\item {\bf SVM}. The standard SVM (BL) is used on the dataset without 
preprocessing 
or modification to weight the class distributions.
\item {\bf Data-level methods (DL)}. The methods in this group are used to 
preprocess 
the data. Then, the edited dataset is used to train an SVM. This group consists of 
ROS~\cite{He2009}, 
SMOTE~\cite{Chawla2002}, SVMSMOTE~\cite{Nguyen2011}, RUS~\cite{He2009}, and 
CNN~\cite{Hart1968}.
\item {\bf Algorithm-level methods (AL)}. This group encompasses 
SVMDC~\cite{Veropoulos1999}, uNBSVM~\cite{Datta2015}, 
WK-SMOTE~\cite{Mathew2018}, CSSVM~\cite{Iranmehr2019}, and 
RBI-LP-SVM~\cite{Datta2019}.
\end{itemize}
}
Reference methods require defining a set of hyper-parameters to use in 
training. Properly selecting hyper-parameters is a crucial step to compare 
classification algorithms~\cite{Bagnall201, Wainer2017}. We adopted the RBF 
kernel because of its effectiveness with SVM. 
For the sake of a fair comparison, the hyper-parameters of each method were 
optimized for each dataset independently by optimizing the BER, computed through 
an internal stratified 5-fold cross-validation 
on the training set. The set of hyper-parameters includes: the $\gamma$ value 
optimized in the range of $\left[2^{-10}, 2^{5} \right]$ for all methods; the 
regularization parameter $C$ in the range of $\left[2^{-5}, 2^{10} \right]$ for 
SVM, data-level methods, and CSSVM; the regularization parameter for positive 
($C^+$) and negative ($C^-$) class optimized between $\left[2^{-5}, 2^{10} 
\right]$ for both WK-SMOTE and SVMDC; the regularization parameter for synthetic 
samples ($C^s$) optimized between $\left[2^{-5}, 2^{10} \right]$ for WK-SMOTE; the 
weight factor for positive ($\omega_p$) and negative ($\omega_n$) class in the 
range of $\left[0, 1\right]$ for uNBSVM; the cost-sensitive parameter 
$\left(\kappa\right)$ ranges from zero to one and the margin violation weight 
$\left(C_1\right)$ is the range of $\left[1,10\right]$ for CSSVM; the amount of 
sampling is searched in the range of $\left[\frac{n_{m}+1}{n_{M}},1\right]$, 
with $n_m$ and $n_M$ as the number of samples in the minority and majority class, 
respectively, for data-level methods and WK-SMOTE, and the number of neighbors is 
between $\left[1,\frac{n_m}{2}\right]$ for SMOTE, SVMSMOTE, and WK-SMOTE. 
SHADE was also used to optimize the hyper-parameters, which ran with a population 
size equals to 30 and the stopping criteria considered performing 1,000 fitness 
function evaluations or that the standard deviation of fitness values of 
the 
population is below $0.001$. 

We provide the implementation of EBCS-SVM, datasets, splits, and detailed results 
in each partition as supplementary material. The supplementary material can be 
downloaded at 
\url{www.cimat.mx/~alejandro.rosales/resources/EBCSSVM.tar.gz}.

\section{Experimental Results and Discussion} \label{sec:experiments_results}

This section presents the experimental results reported by EBCS-SVM and 
reference methods. Section~\ref{sec:small_ir} shows the results obtained with 
datasets with small IR. Next, Section~\ref{sec:medium_ir} considers the 35 
datasets with medium IR, and Section~\ref{sec:high_ir} presents the results 
using the 25 datasets with high IR. Finally, Section~\ref{sec:computational_time} 
compares the training time required by each method.

\subsection{Experiments on Small Imbalance Ratio} \label{sec:small_ir}

\begin{figure*}[!ht]
\centering
\subfloat[SVM -- EBC-SVM]{
\includegraphics[width=0.23\textwidth]{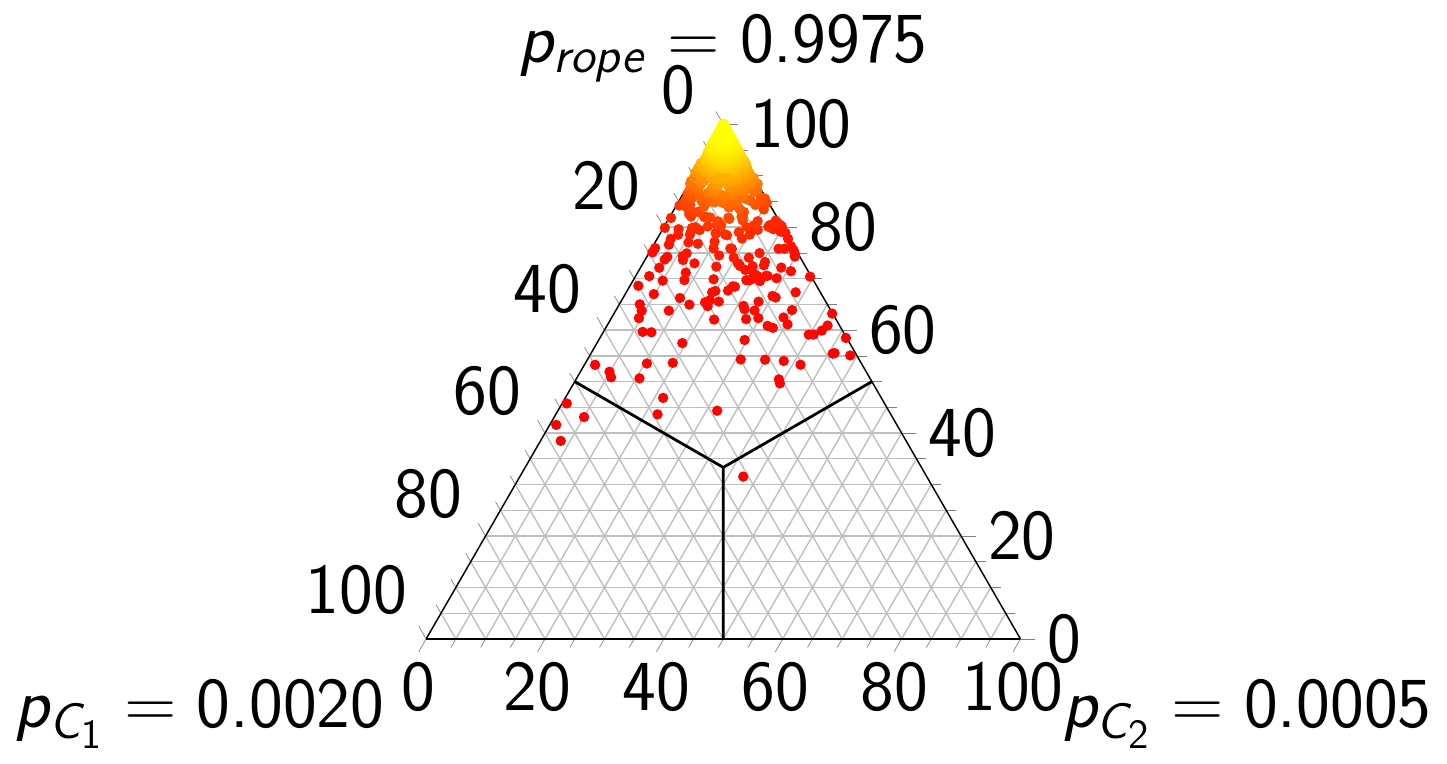}
\label{fig:small_svm_bar}}
\subfloat[ROS -- EBC-SVM]{
\includegraphics[width=0.23\textwidth]{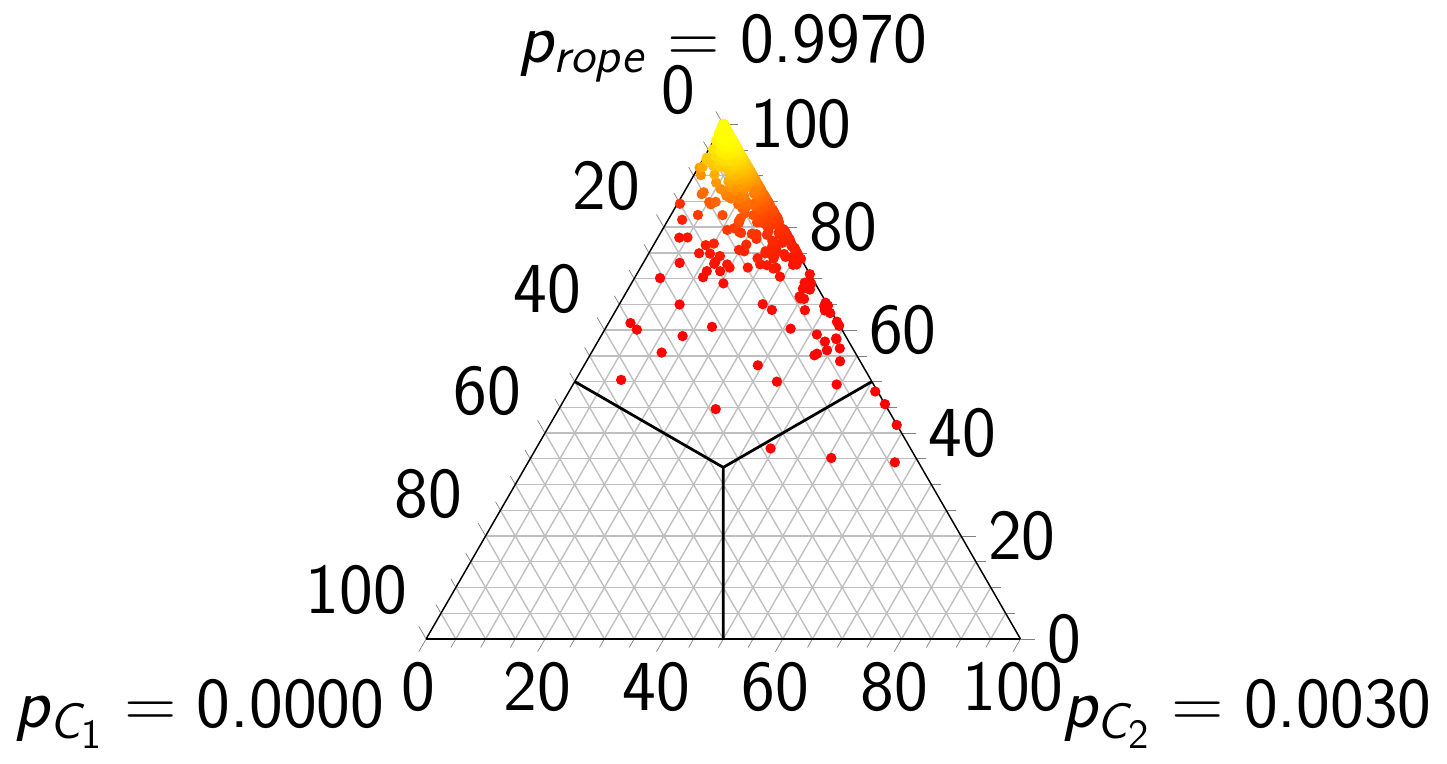}
\label{fig:small_ros_bar}}
\subfloat[SMOTE -- EBC-SVM]{
\includegraphics[width=0.23\textwidth]{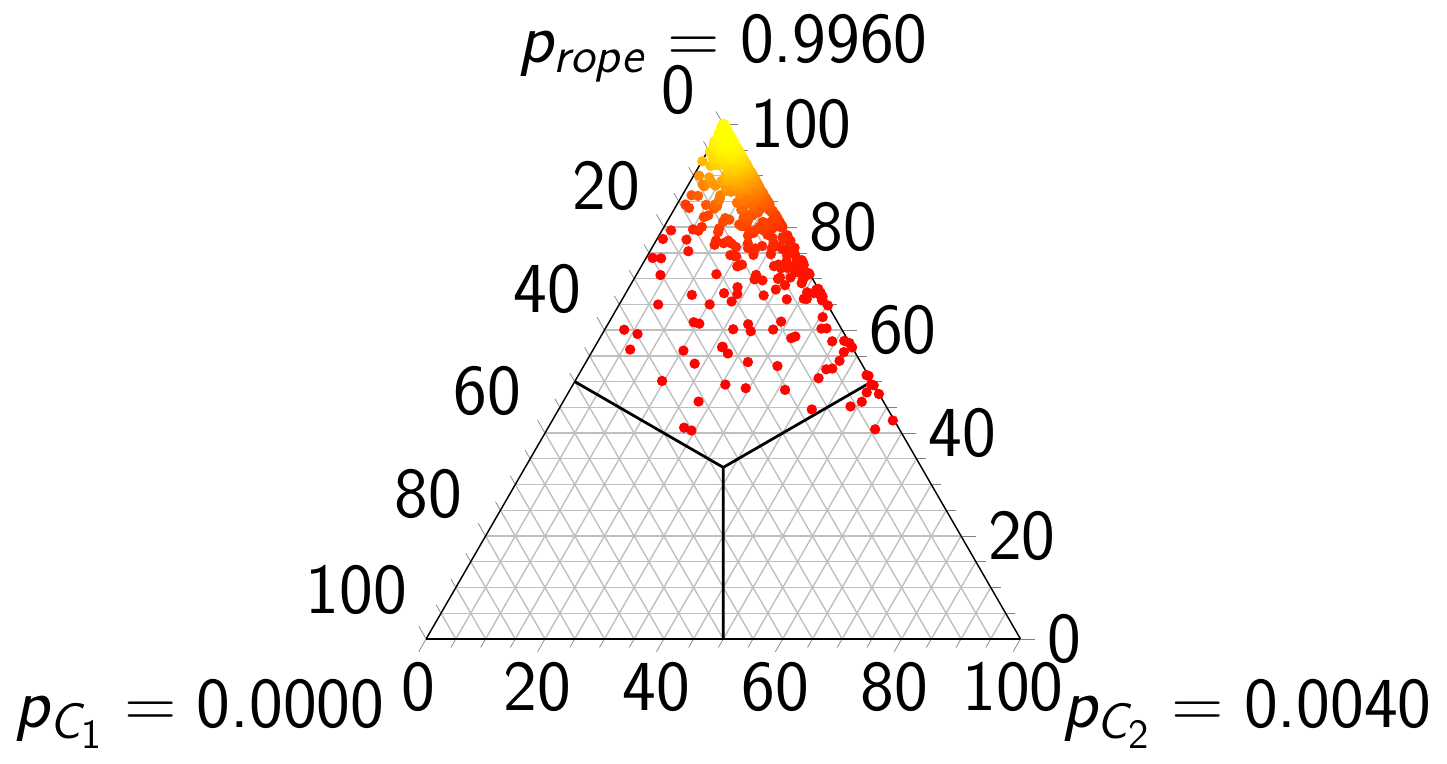}
\label{fig:small_smote_bar}}
\subfloat[SVMSMOTE -- EBC-SVM]{
\includegraphics[width=0.23\textwidth]{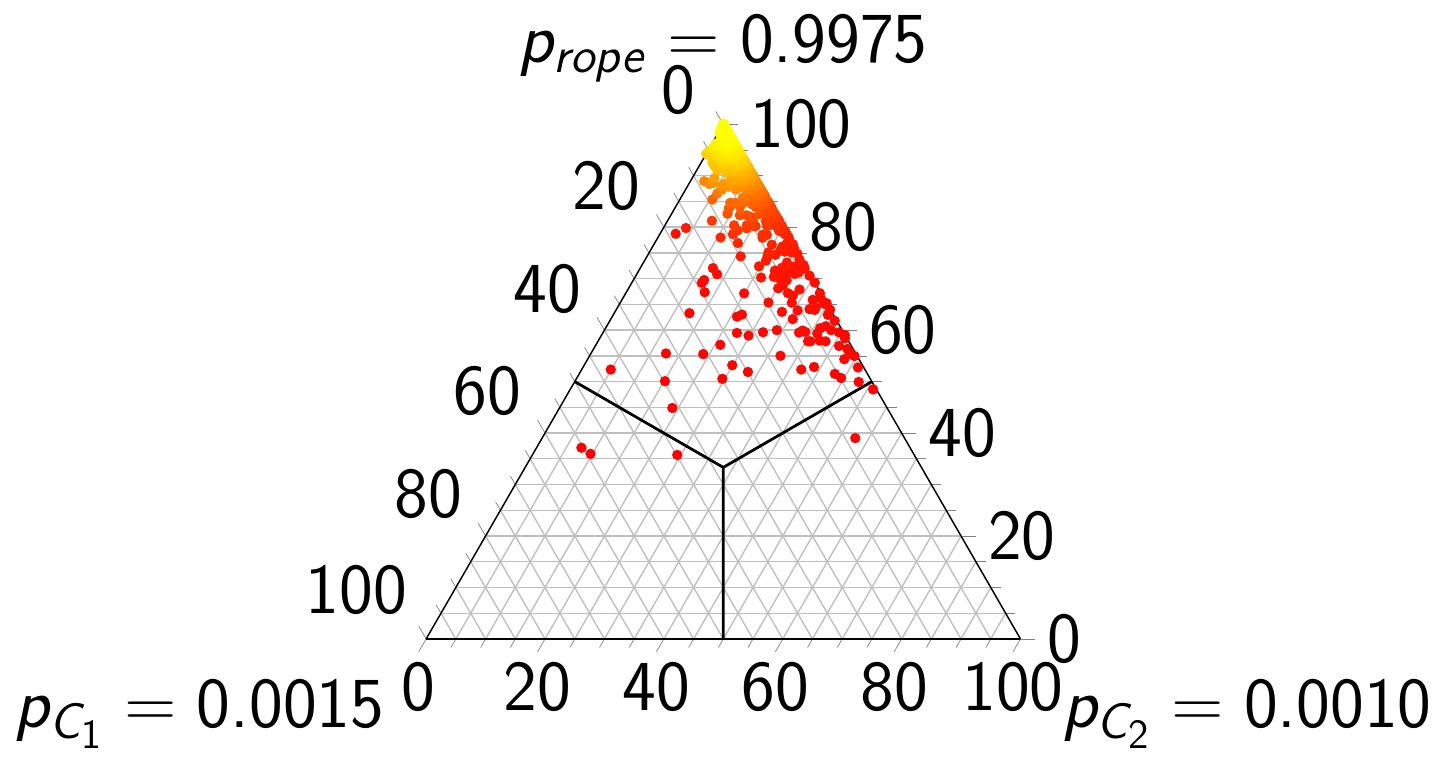}
\label{fig:small_svmsmote_bar}}
\qquad
\subfloat[RUS -- EBC-SVM]{
\includegraphics[width=0.23\textwidth]{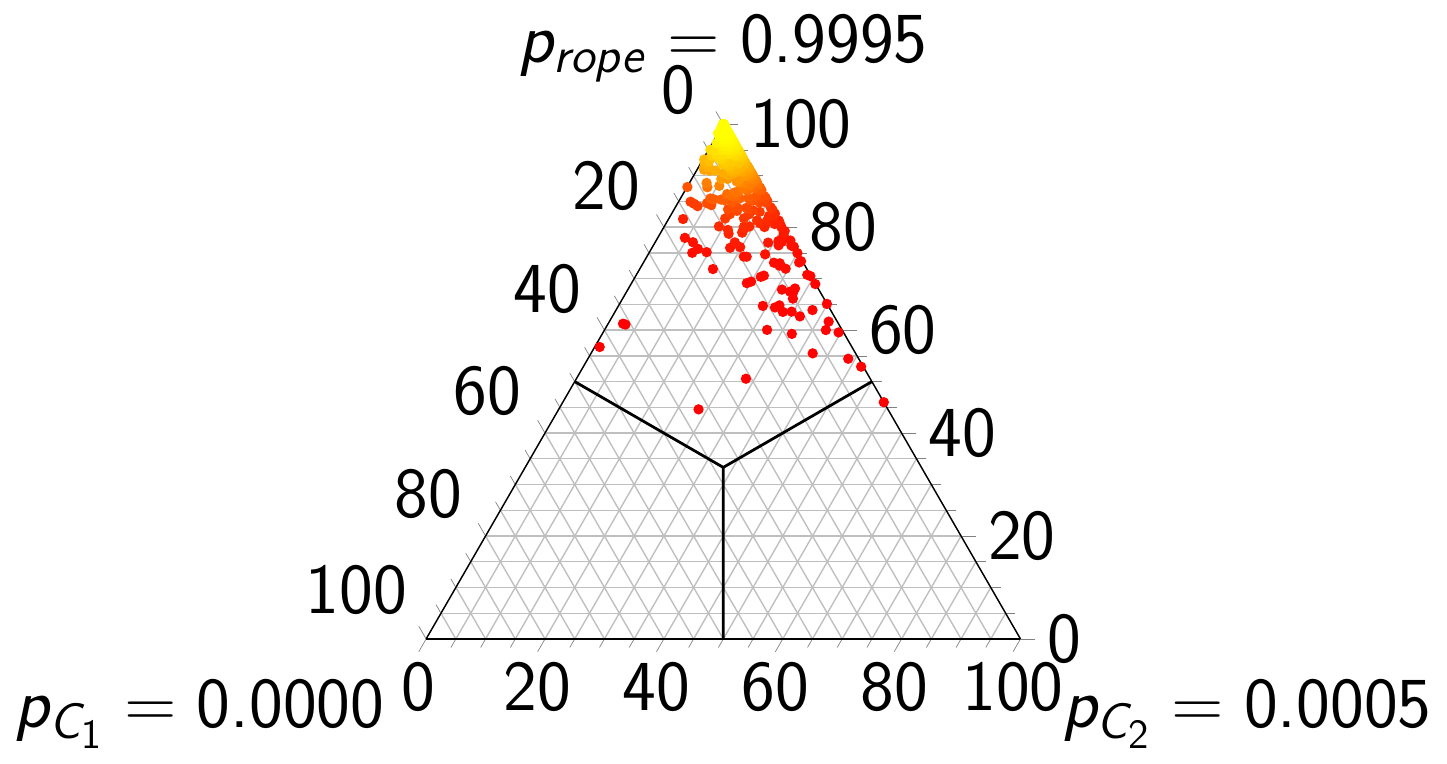}
\label{fig:small_rus_bar}}
\subfloat[CNN -- EBC-SVM]{
\includegraphics[width=0.23\textwidth]{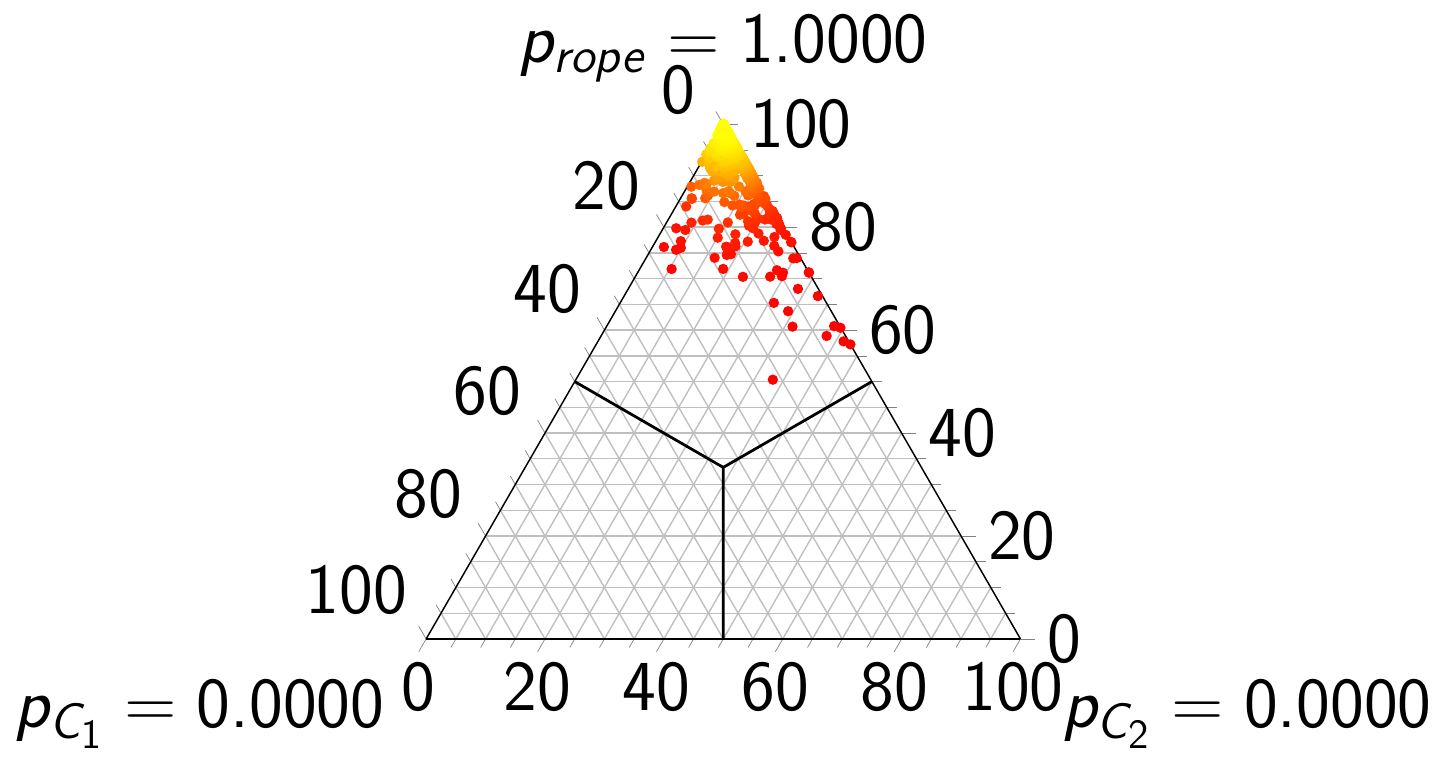}
\label{fig:small_cnn_bar}}
\subfloat[SVMDC -- EBC-SVM]{
\includegraphics[width=0.23\textwidth]{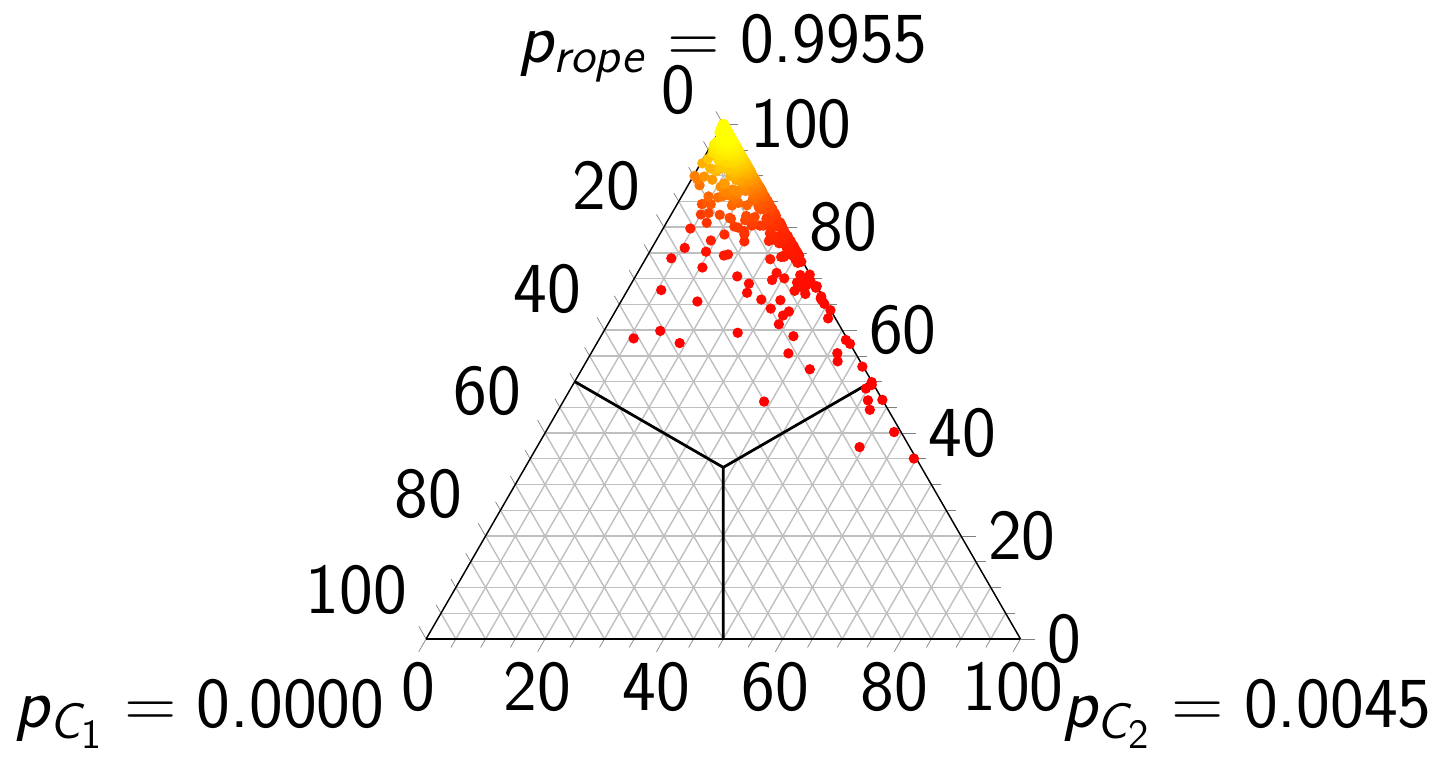}
\label{fig:small_svmdc_bar}}
\subfloat[CSSVM -- EBC-SVM]{
\includegraphics[width=0.23\textwidth]{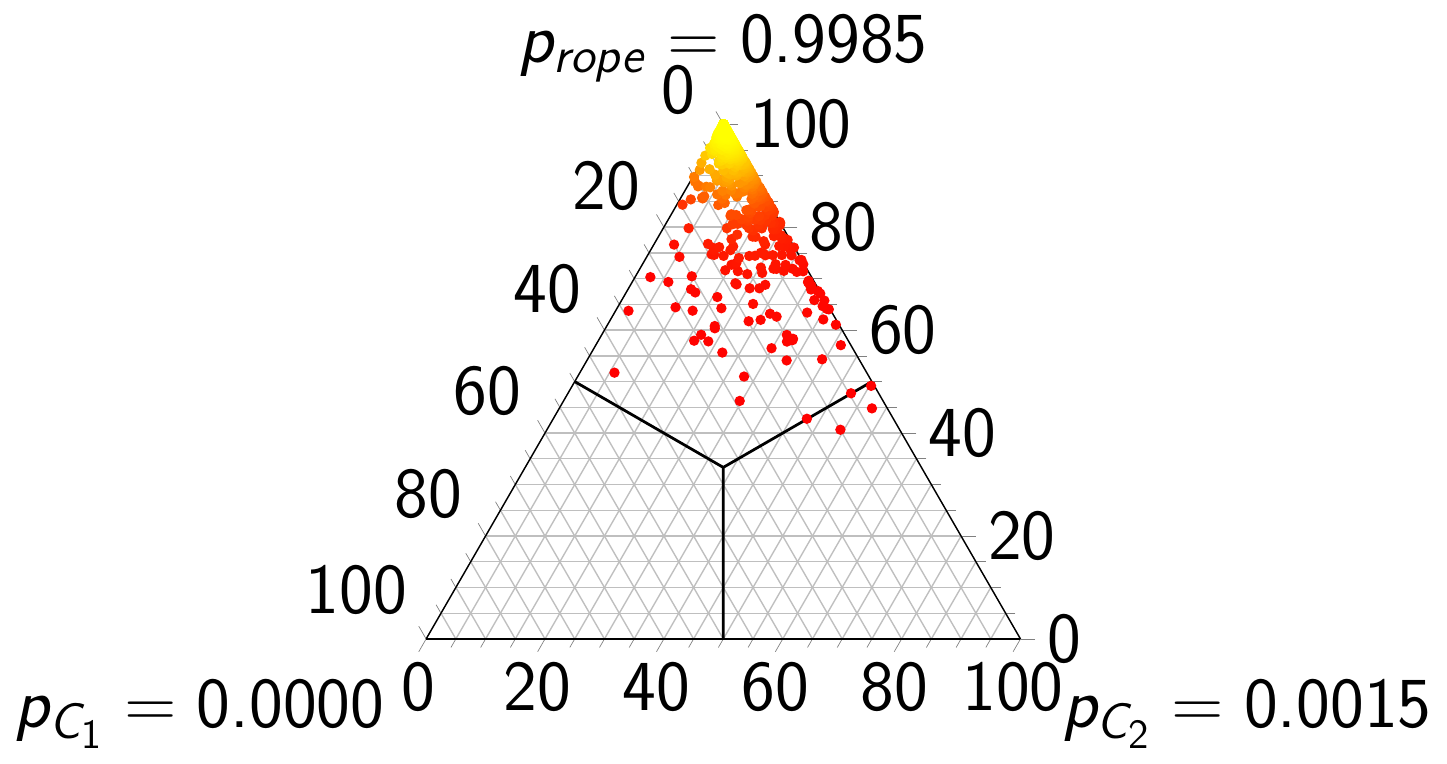}
\label{fig:small_cssvm_bar}}
\qquad
\subfloat[uNBSVM]{
\includegraphics[width=0.23\textwidth]{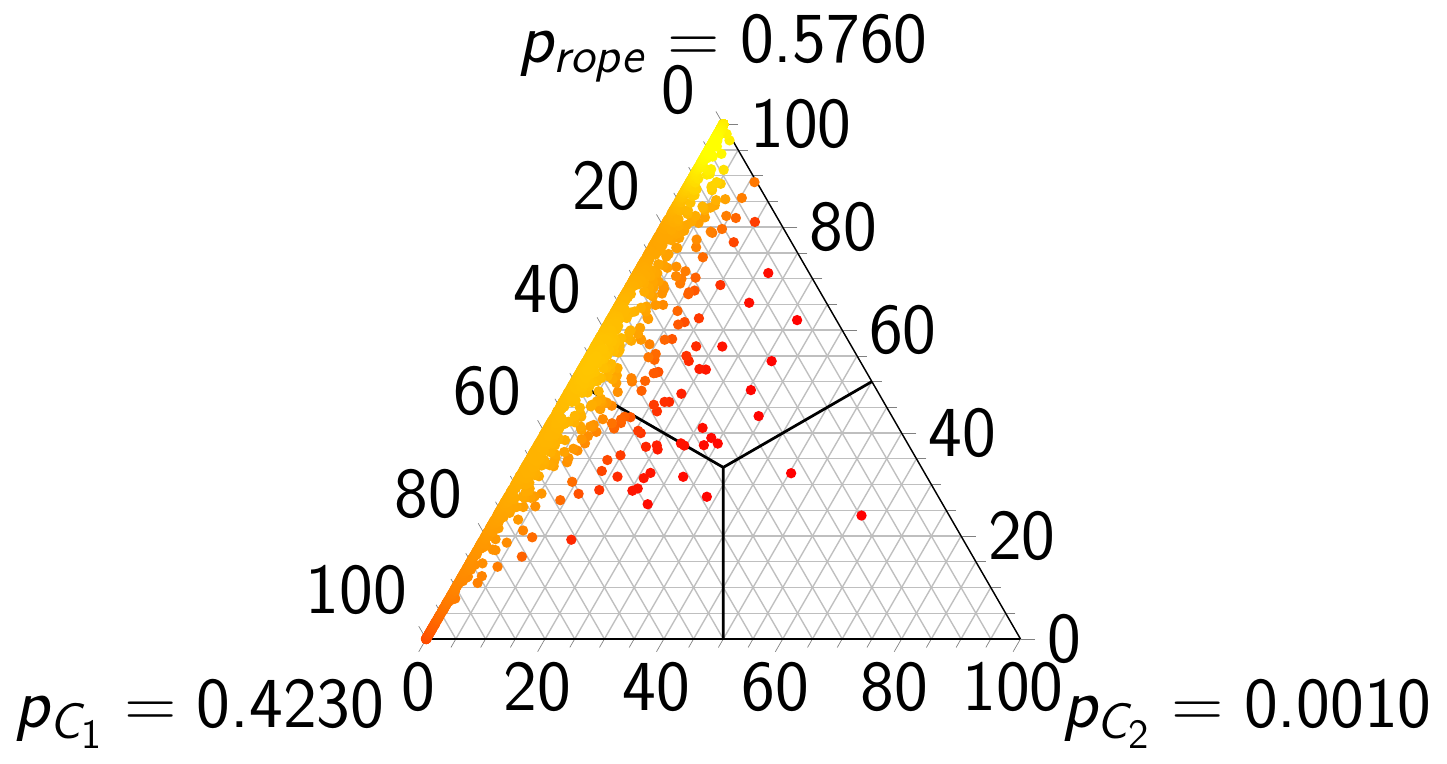}
\label{fig:small_unbsvm_bar}}
\subfloat[WK-SMOTE -- EBC-SVM]{
\includegraphics[width=0.23\textwidth]{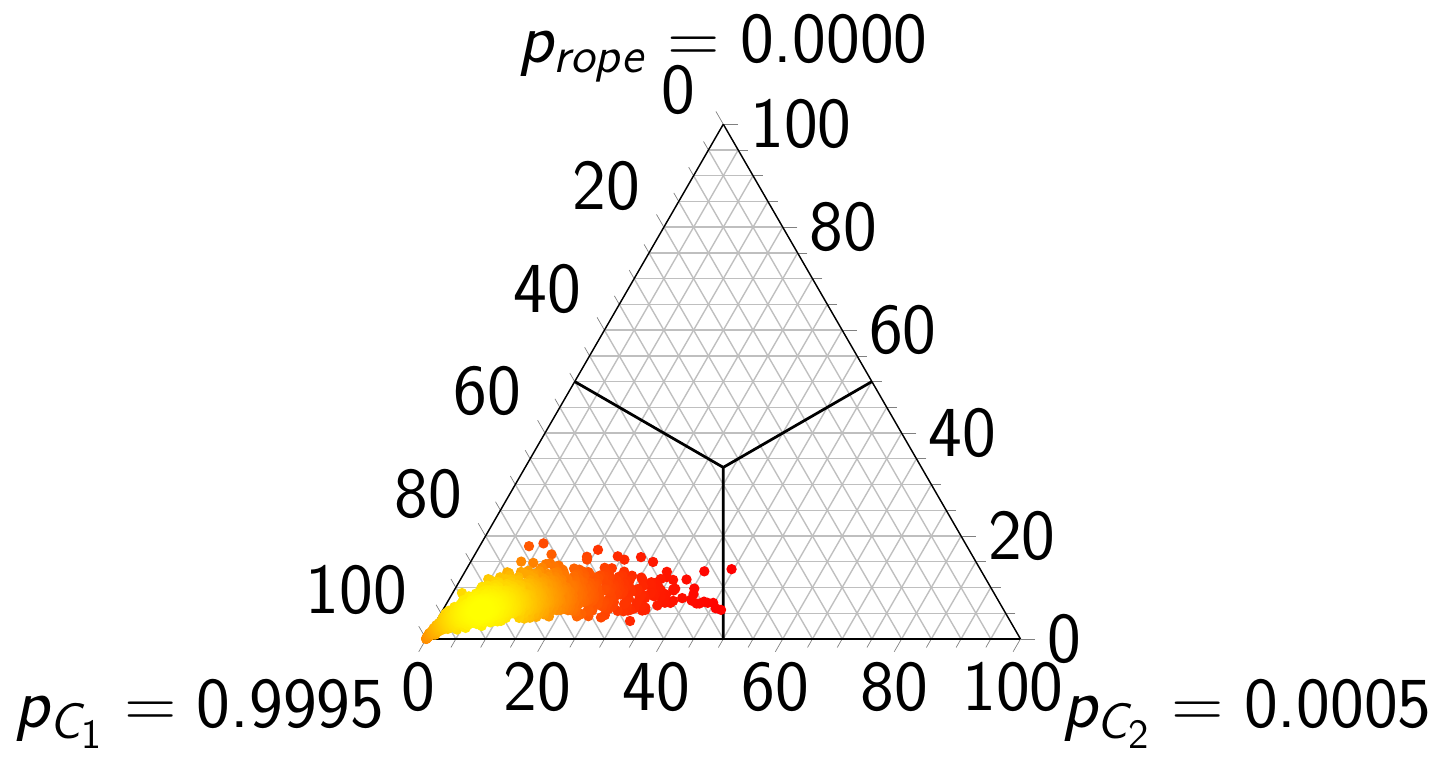}
\label{fig:small_wksmote_bar}}
\subfloat[RBI-LP-SVM -- EBC-SVM]{
\includegraphics[width=0.23\textwidth]{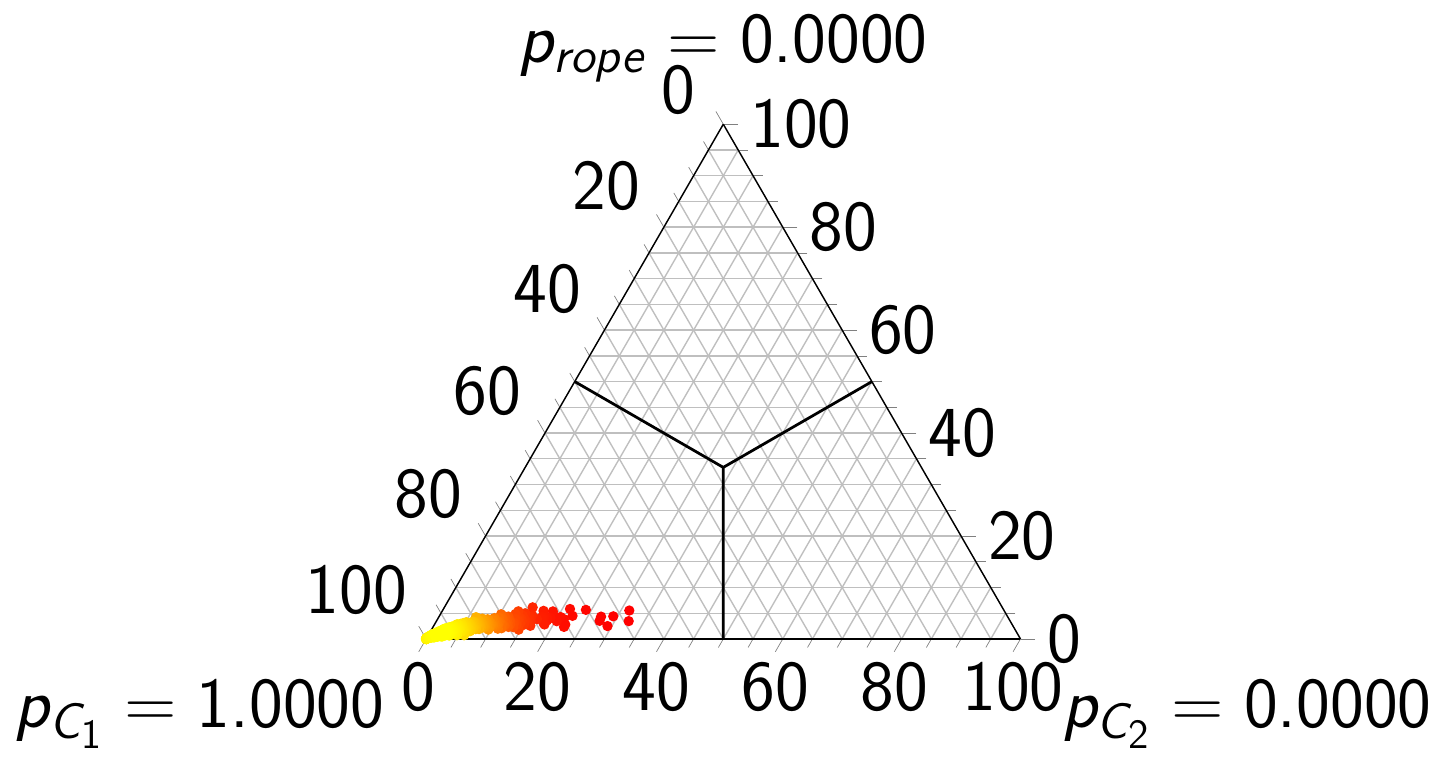}
\label{fig:small_rbi_bar}}
\caption{Posterior distribution for BAR when comparing EBCS-SVM with 
the reference methods on small IR datasets. The region of the bottom-left 
represents EBCS-SVM, the region at the top is for rope, and the region in the 
bottom-right represents the reference method.}
\label{fig:small_posterior}
\end{figure*}

In this section, we focused on comparing the performance of EBCS-SVM against 
reference methods. Table~\ref{tab:results_small_ir} shows the results on the ten 
datasets with a small IR. The reported results correspond to the average and 
standard deviations for BAR, BMI, GM, uF1, and uMCC scores. 
Fig.~\ref{fig:small_posterior} graphically depicts barycenter plots for the 
posterior probabilities reported by the hierarchical Bayesian test for the BAR 
score. Each point on the barycenter plot represents an estimate of the probability 
that it belongs to each region. Based on the results and the analysis carried out 
by the hierarchical Bayesian test, the following observations are highlighted:

\begin{table*}
\centering
 \caption{Obtained results on small IR datasets for BAR, BMI, GM, uF1, and 
uMCC scores.}
\label{tab:results_small_ir}
\begin{tabular}{llrrrrr}
\hline
Fam. & Method & BAR & BMI & GM & uF1 & uMCC\\
 \hline
BL & SVM & $0.8626 \pm 0.1461$ & $0.7251 \pm 0.2922$ & $0.8494 \pm 0.1651$ & 
$0.8312 
\pm 0.1904$ & $0.7343 \pm 0.2831$ \\
DL & ROS & $0.8726 \pm 0.1321$ & $0.7452 \pm 0.2643$ & $0.8643 \pm 0.1430$ & 
$0.8515 
\pm 0.1600$ & $0.7528 \pm 0.2554$ \\
DL & SMOTE & $0.8719 \pm 0.1328$ & $0.7439 \pm 0.2656$ & $0.8632 \pm 0.1447$ & 
$0.8502 \pm 0.1628$ & $0.7512 \pm 0.2569$ \\
DL & SVMSMOTE & $0.8723 \pm 0.1345$ & $0.7447 \pm 0.2690$ & $0.8627 \pm 0.1487$ & 
$0.8499 \pm 0.1681$ & $0.7519 \pm 0.2601$ \\
DL & RUS & $0.8712 \pm 0.1335$ & $0.7424 \pm 0.2670$ & $0.8636 \pm 0.1442$ & 
$0.8548 
\pm 0.1560$ & $0.7489 \pm 0.2593$ \\
DL & CNN & $0.8688 \pm 0.1345$ & $0.7377 \pm 0.2690$ & $0.8616 \pm 0.1448$ & 
$0.8586 
\pm 0.1506$ & $0.7441 \pm 0.2613$ \\
AL & SVMDC & $0.8730 \pm 0.1324$ & $0.7460 \pm 0.2649$ & $0.8644 \pm 0.1446$ & 
$0.8552 \pm 0.1570$ & $0.7529 \pm 0.2569$ \\
AL & CSSVM & $\mathbf{0.8740 \pm 0.1267}$ & $\mathbf{0.7480 \pm 0.2533}$ & 
$\mathbf{0.8689 \pm 0.1327}$ & $\mathbf{0.8674 \pm 0.1357}$ & $\mathbf{0.7535 
\pm 0.2481}$ \\
AL & WK-SMOTE & $0.7322 \pm 0.1137$ & $0.4643 \pm 0.2275$ & $0.6194 \pm 0.1743$ & 
$0.5753 \pm 0.1930$ & $0.5069 \pm 0.2181$ \\
AL & uNBSVM & $0.8419 \pm 0.1325$ & $0.6839 \pm 0.2650$ & $0.8257 \pm 0.1479$ & 
$0.8415 \pm 0.1310$ & $0.6960 \pm 0.2582$ \\
AL & RBI-LP-SVM & $0.5293 \pm 0.0568$ & $0.0587 \pm 0.1136$ & $0.0676 \pm 0.1136$ 
& 
$0.6724 \pm 0.0474$ & $0.0604 \pm 0.1135$ \\
AL & EBCS-SVM & $0.8696 \pm 0.1310$ & $0.7391 \pm 0.2619$ & $0.8585 \pm 0.1473$ & 
$0.8622 \pm 0.1419$ & $0.7468 \pm 0.2536$ \\
 \hline
\end{tabular}
\end{table*}

\begin{itemize}
\item Most methods showed a competitive performance when dealing with small IR. 
\item For metrics BAR, GM, and uF1, most methods reported scores above 0.820. The 
exceptions were WK-SMOTE and RBI-LP-SVM, which obtained performances below 0.800.
On the other hand, for metrics BMI and MCC, most methods reported scores above 
0.710, except for WK-SMOTE, nNBSVM, and RBI-LP-SVM.
\item CSSVM obtained the highest performance in all metrics. SVMDC was the 
second-best position for BAR, BMI, GM, and uMCC, and the third-best for uF1. On the 
other hand, EBCS-SVM ranked second-best for uF1; it was in the seventh position for 
BAR, BMI, and uMCC; and in the eighth position for GM.
\item The hierarchical Bayesian tests provided strong evidence on the 
practical equivalence between CSSVM and EBC-SVM. We can observe similar behavior 
when EBCS-SVM is compared with SVM, ROS, SMOTE, SVMSMOTE, RUS, CNN, and SVMDC. 
Thus, EBCS-SVM exhibited a performance practically similar to that of CSSVM, the 
best-ranked method. 
\item Among data-level methods, ROS had the best average performance over the 
ten datasets with a small IR for BAR, BMI, GM, and uMCC, while for uF1 CNN was 
the best data-level approach. Conversely, CNN was the worst data-level 
method for BAR, DMI, GM, and uMCC, while SVMSMOTE was the worst for uF1.
\item RBI-LP-SVM showed the lowest performance among algorithm-level methods. 
The hierarchical Bayesian test reported probabilities above 0.999 in 
the region of EBCS-SVM for all metrics.
\item Algorithm-level methods generally reported better performance than 
data-level methods.
\end{itemize}

EBCS-SVM, uNBSVM, CSSVM, and ROS were highly effective methods to classify 
problems with a small IR. In the next section, we delved into our analysis when 
the IR increases.

\subsection{Experiments with Medium Imbalance Ratio} \label{sec:medium_ir}

\begin{figure*}[!ht]
\centering
\subfloat[SVM -- EBC-SVM]{
\includegraphics[width=0.23\textwidth]{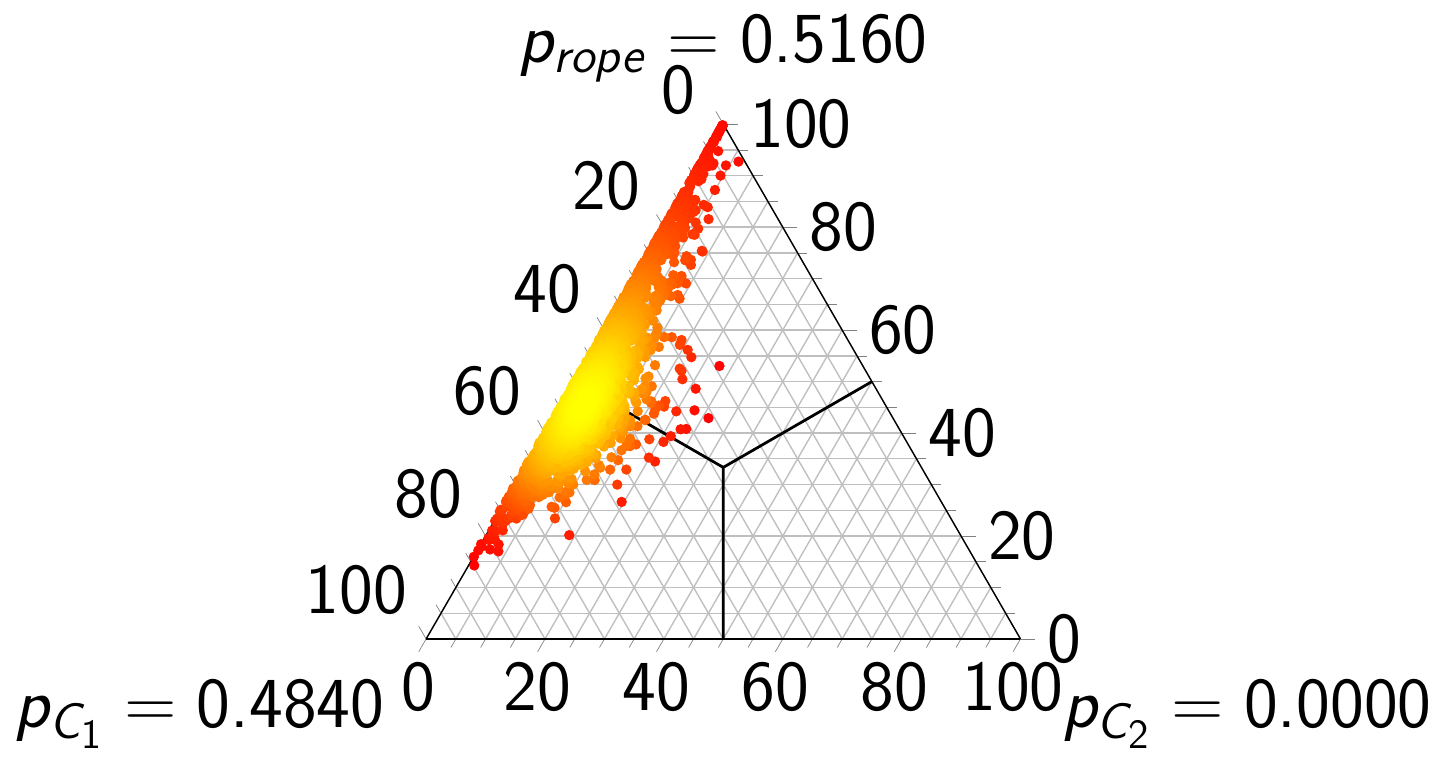}
\label{fig:medium_svm_bar}}
\subfloat[ROS -- EBC-SVM]{
\includegraphics[width=0.23\textwidth]{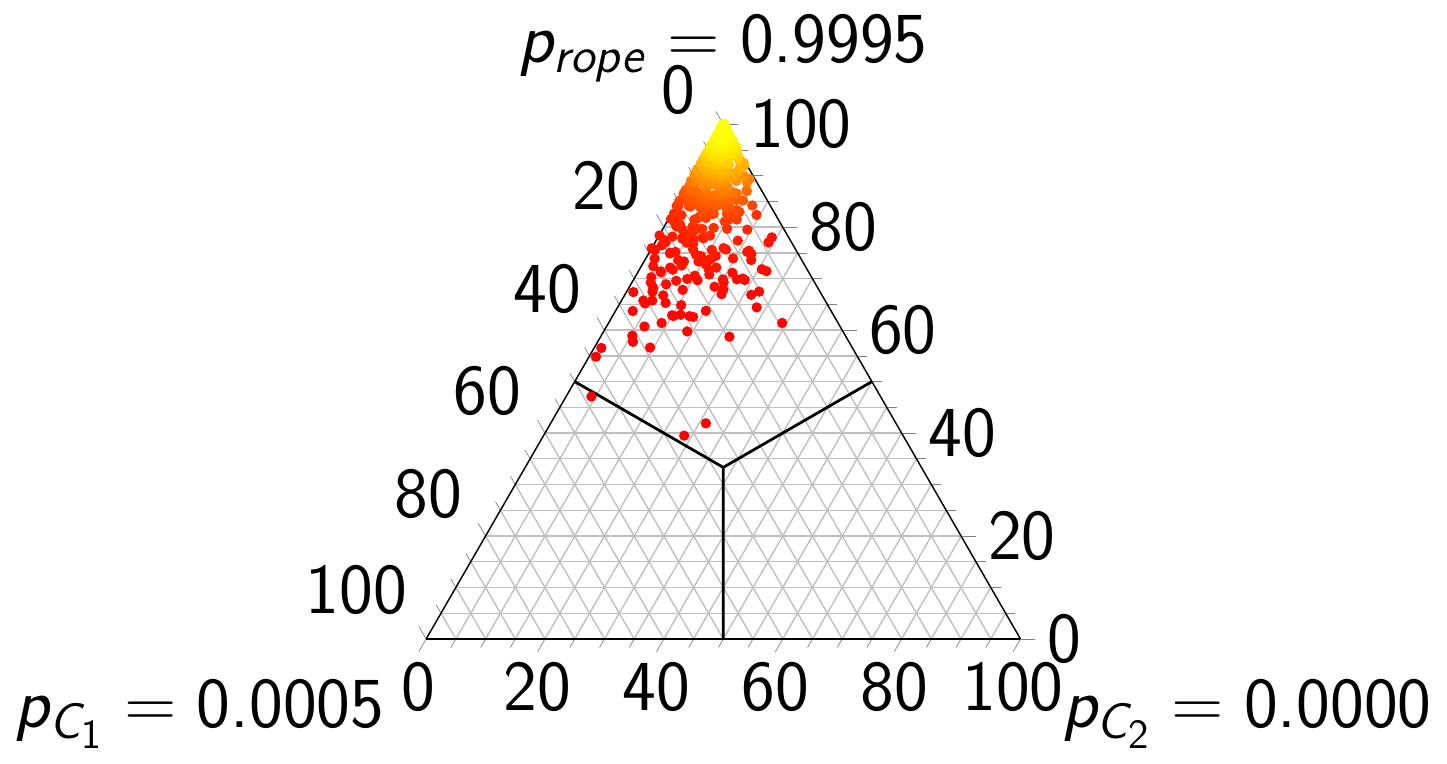}
\label{fig:medium_ros_bar}}
\subfloat[SMOTE -- EBC-SVM]{
\includegraphics[width=0.23\textwidth]{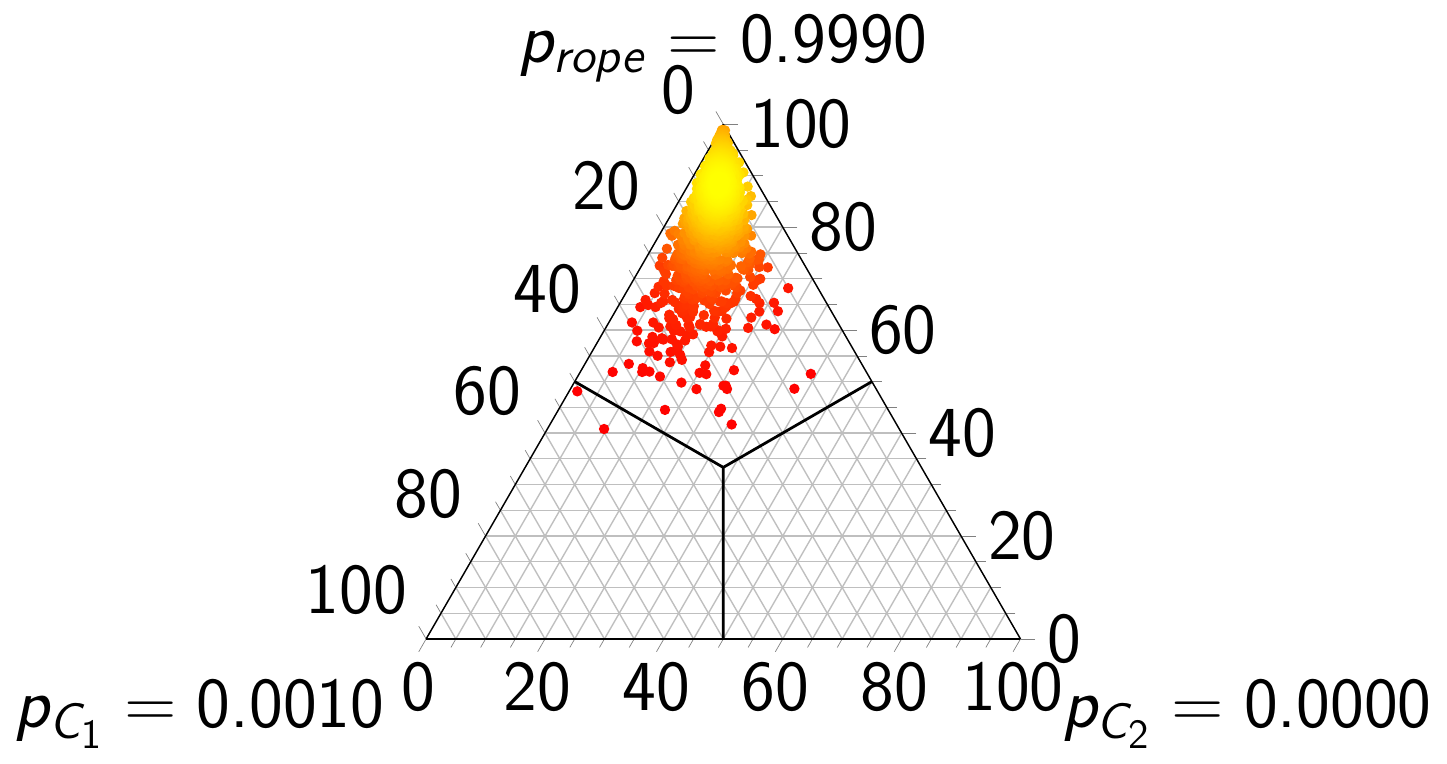}
\label{fig:medium_smote_bar}}
\subfloat[SVMSMOTE -- EBC-SVM]{
\includegraphics[width=0.23\textwidth]{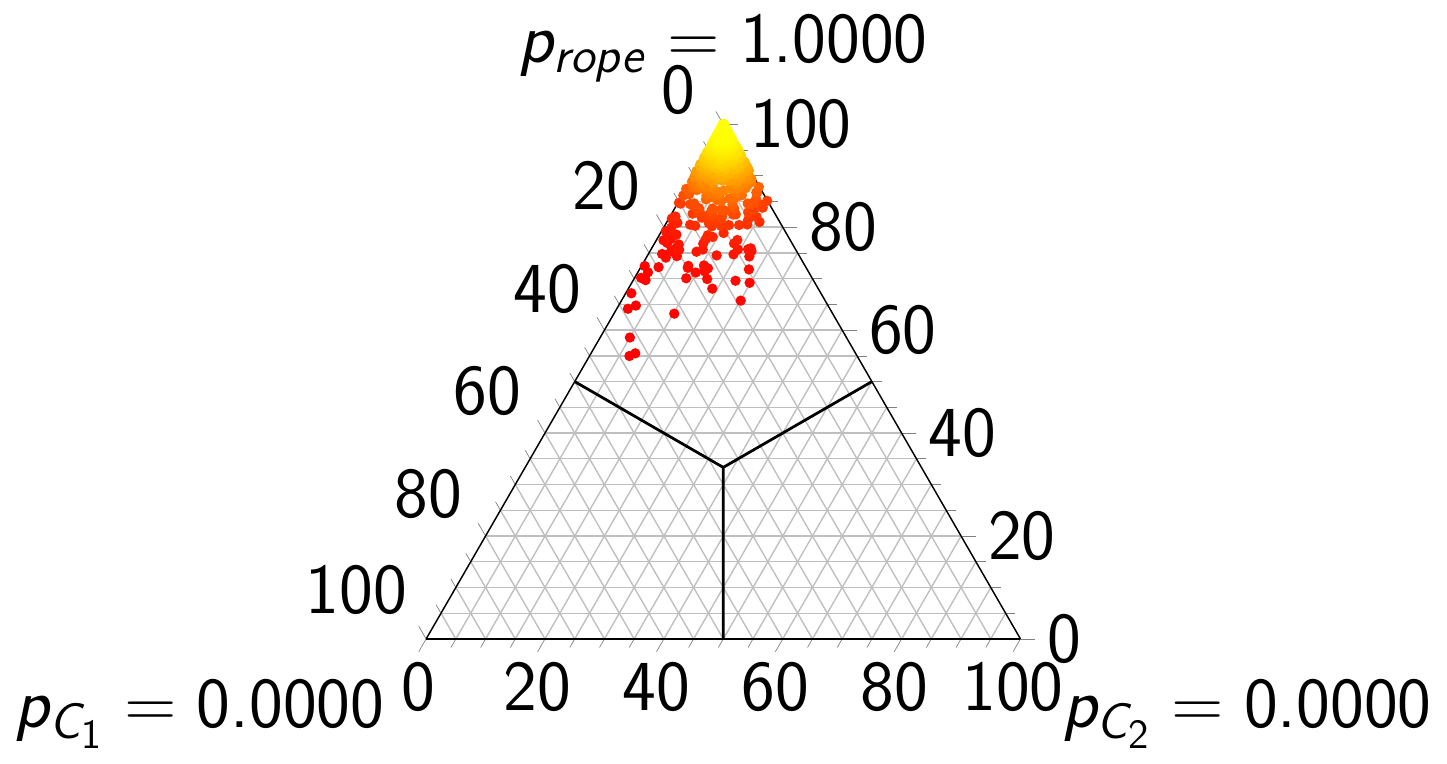}
\label{fig:medium_svmsmote_bar}}
\qquad
\subfloat[RUS -- EBC-SVM]{
\includegraphics[width=0.23\textwidth]{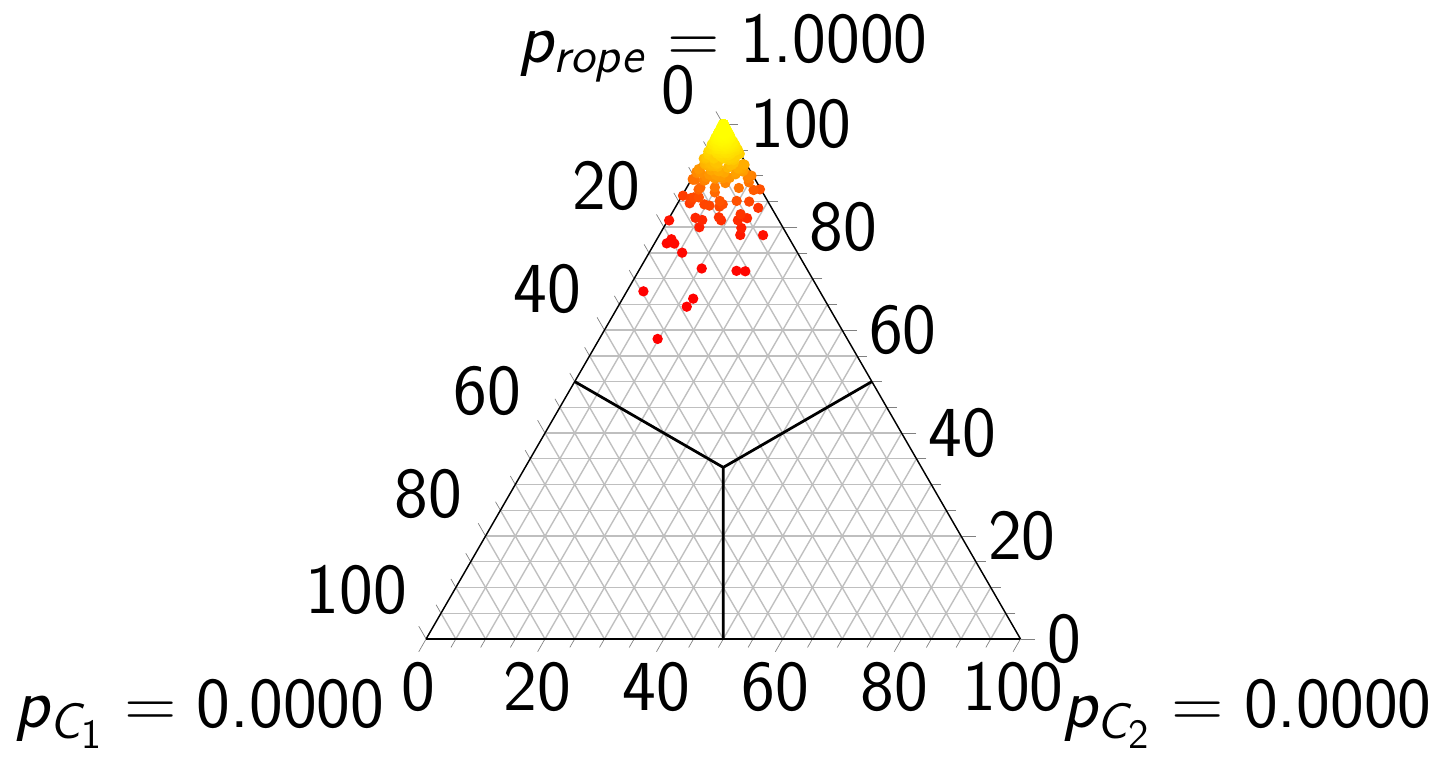}
\label{fig:medium_rus_bar}}
\subfloat[CNN -- EBC-SVM]{
\includegraphics[width=0.23\textwidth]{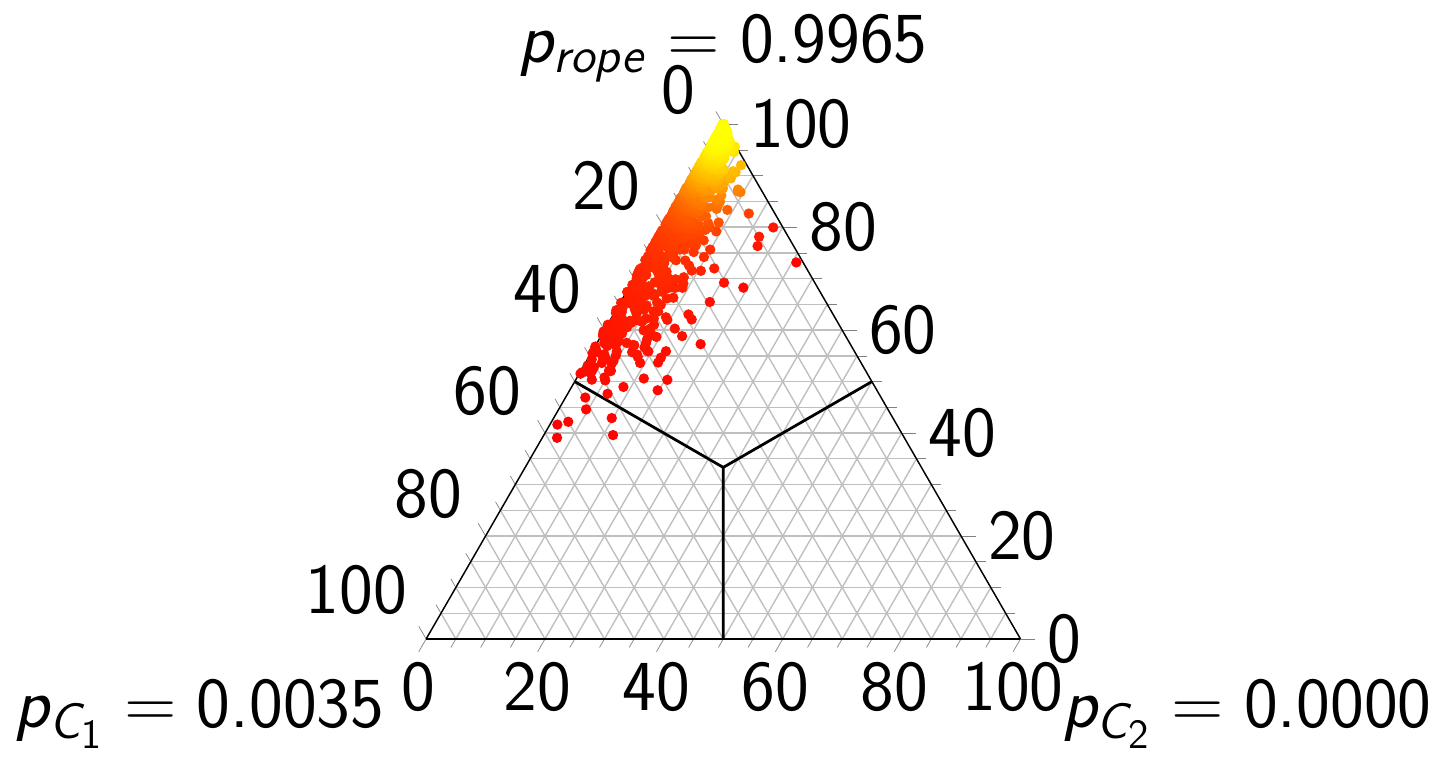}
\label{fig:medium_cnn_bar}}
\subfloat[SVMDC -- EBC-SVM]{
\includegraphics[width=0.23\textwidth]{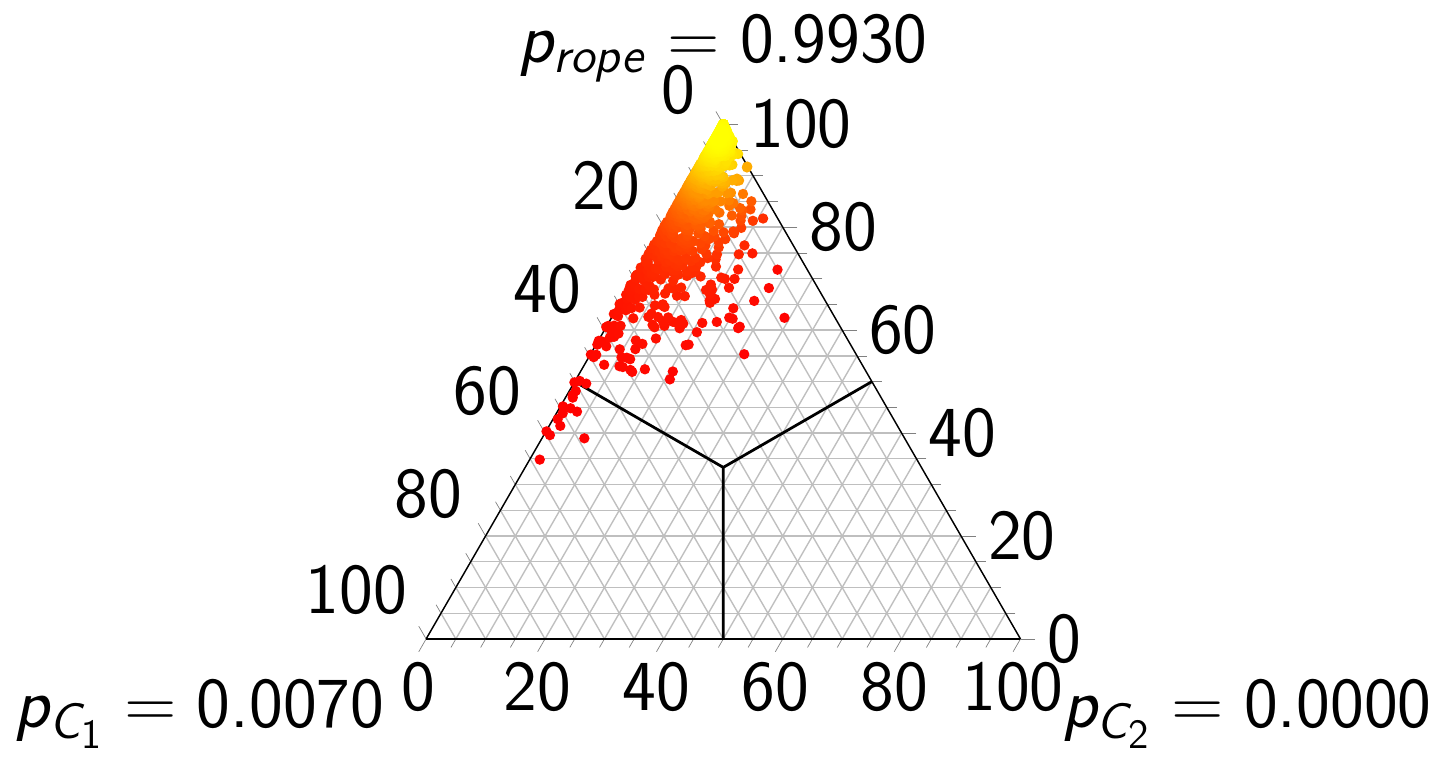}
\label{fig:medium_svmdc_bar}}
\subfloat[CSSVM -- EBC-SVM]{
\includegraphics[width=0.23\textwidth]{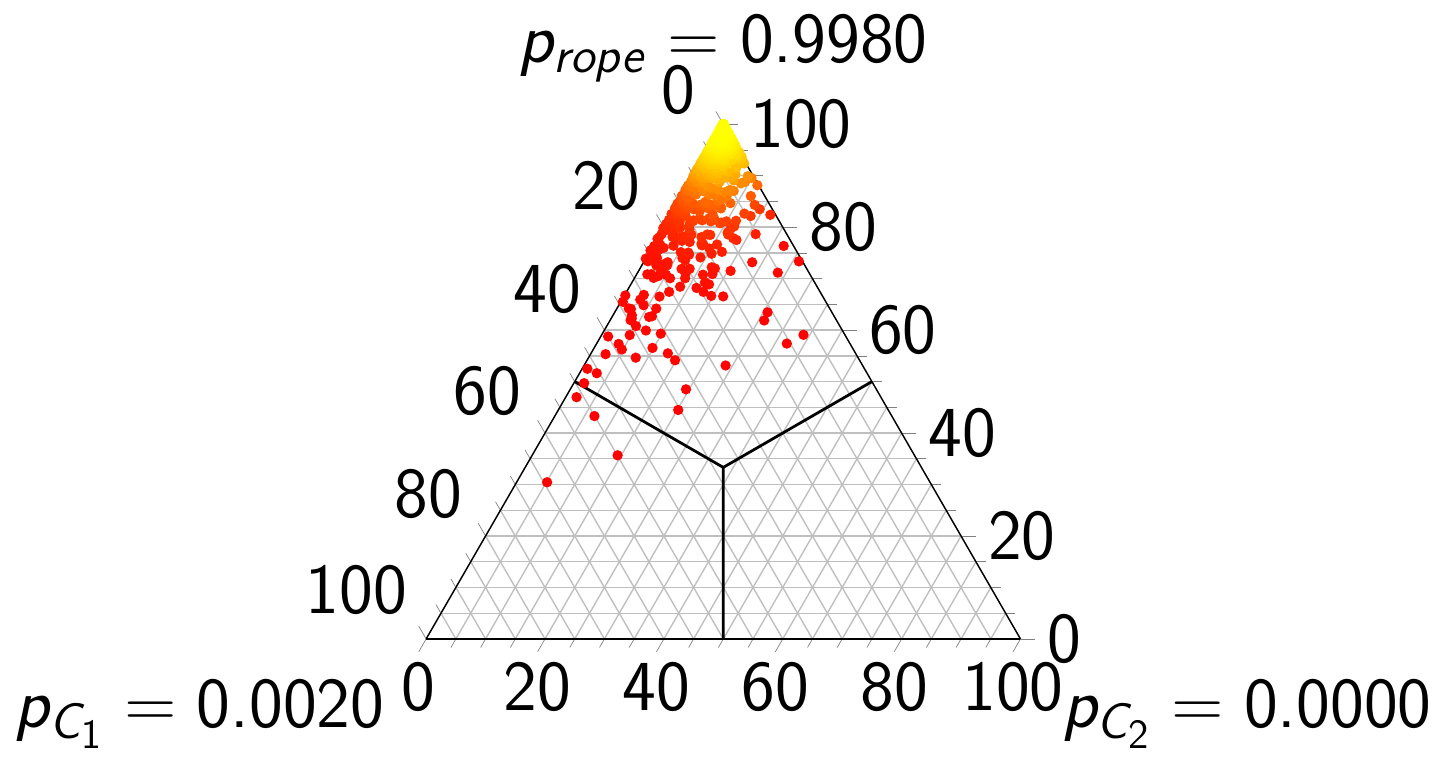}
\label{fig:medium_cssvm_bar}}
\qquad
\subfloat[uNBSVM]{
\includegraphics[width=0.23\textwidth]{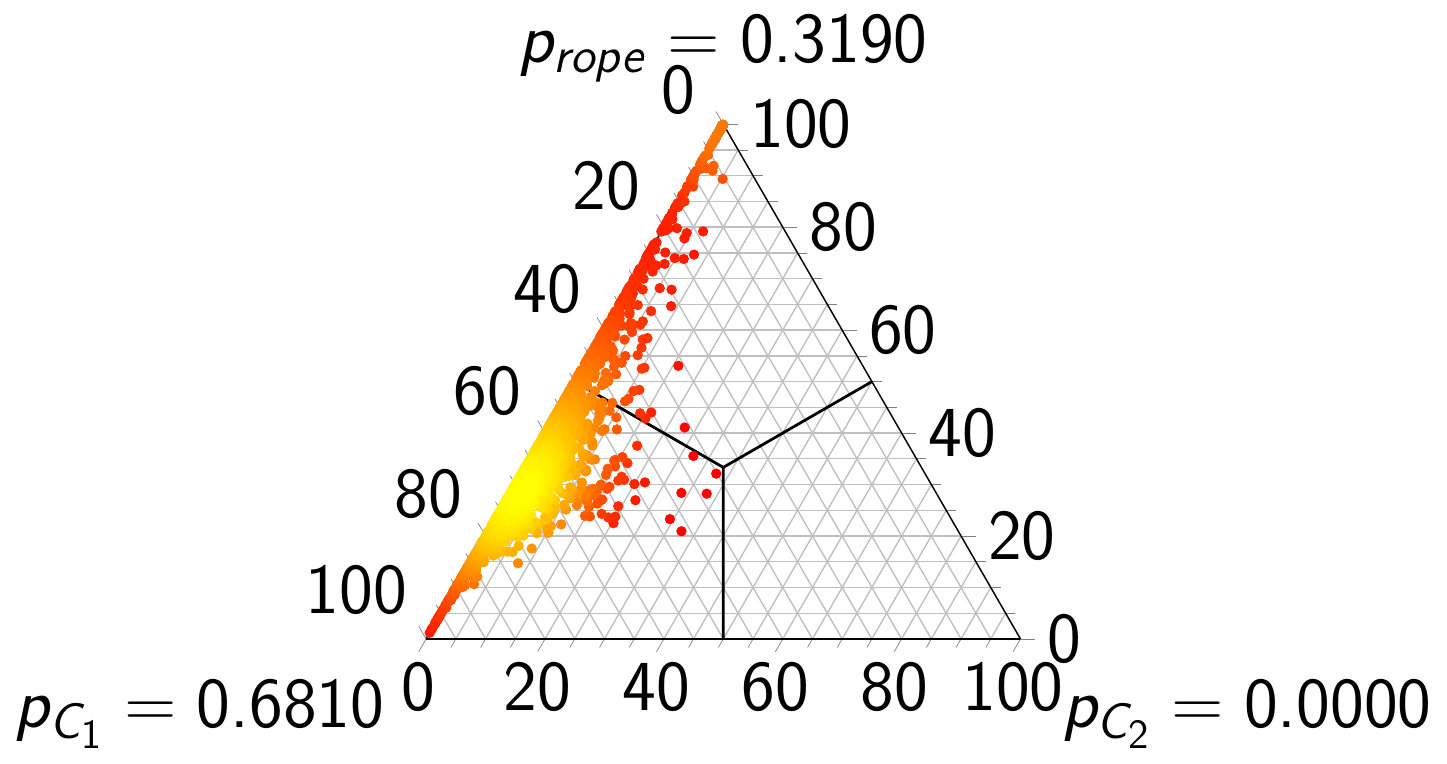}
\label{fig:medium_unbsvm_bar}}
\subfloat[WK-SMOTE -- EBC-SVM]{
\includegraphics[width=0.23\textwidth]{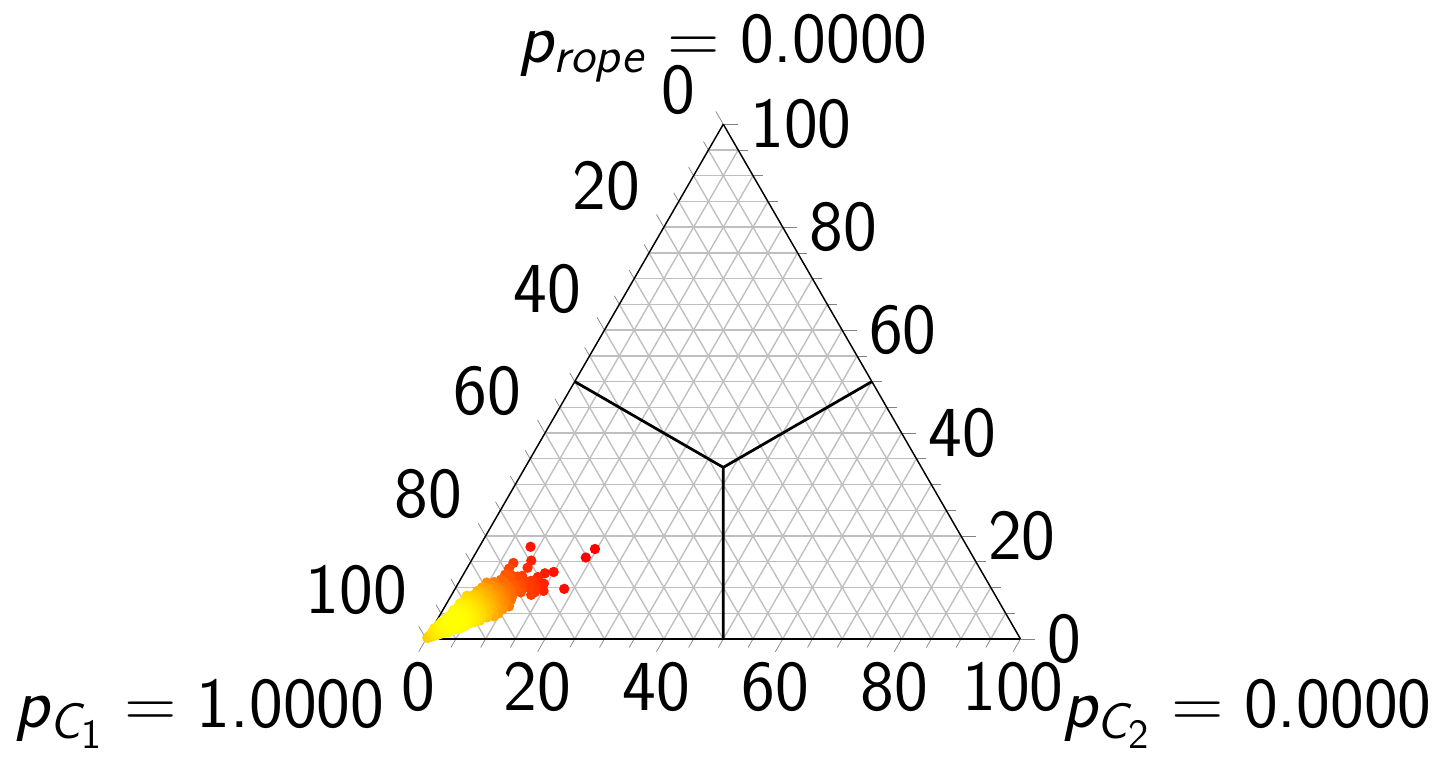}
\label{fig:medium_wksmote_bar}}
\subfloat[RBI-LP-SVM -- EBC-SVM]{
\includegraphics[width=0.23\textwidth]{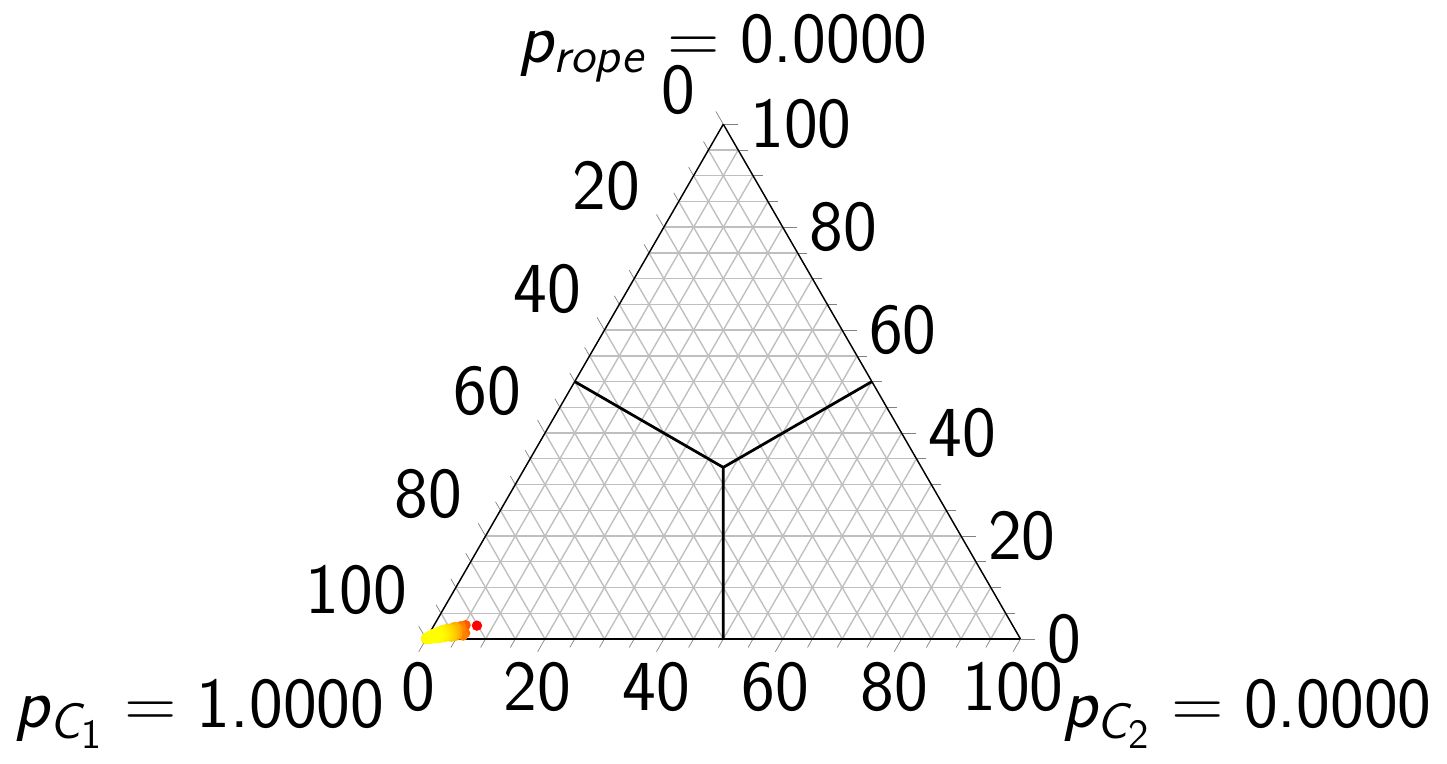}
\label{fig:medium_rbi_bar}}
\caption{Posterior distribution for BAR when comparing EBCS-SVM with 
the reference methods on medium IR datasetss. The region of the bottom-left 
represents EBCS-SVM, the region at the top is for rope, and the region in the 
bottom-right represents the reference method.}
\label{fig:medium_posterior}
\end{figure*}

In this section, we analyzed the performance of EBCS-SVM and reference 
methods when using the 35 benchmark datasets with medium IR. 
Table~\ref{tab:results_medium_ir} presents the average results obtained by each 
method, and Fig.~\ref{fig:medium_posterior} shows the posterior plots for the 
BAR score when EBCS-SVM is compared with reference methods. From these, the 
following is pointed out: 

\begin{table*}
\centering
 \caption{Obtained results on medium IR datasets for BAR, BMI, GM, uF1, and 
uMCC scores.}
\label{tab:results_medium_ir}
\begin{tabular}{llrrrrr}
\hline
 Fam. & Method & BAR & BMI & GM & uF1 & uMCC\\
 \hline
BL & SVM & $0.8420 \pm 0.1226$ & $0.6841 \pm 0.2453$ & $0.7931 \pm 0.1929$ & 
$0.7724 
\pm 0.2107$ & $0.7090 \pm 0.2314$ \\
DL & ROS & $\mathbf{0.8832 \pm 0.0728}$ & $\mathbf{0.7664 \pm 0.1457}$ & 
$\mathbf{0.8693 \pm 0.0808}$ & $0.8579 \pm 0.0892$ & $\mathbf{0.7813 \pm 
0.1395}$ \\
DL & SMOTE & $0.8738 \pm 0.0974$ & $0.7475 \pm 0.1948$ & $0.8455 \pm 0.1656$ & 
$0.8345 \pm 0.1687$ & $0.7613 \pm 0.1921$ \\
DL & SVMSMOTE & $0.8678 \pm 0.0942$ & $0.7356 \pm 0.1884$ & $0.8329 \pm 0.1549$ & 
$0.8216 \pm 0.1599$ & $0.7496 \pm 0.1845$ \\
DL & RUS & $0.8793 \pm 0.0858$ & $0.7586 \pm 0.1716$ & $0.8652 \pm 0.1025$ & 
$\mathbf{0.8603 \pm 0.1073}$ & $0.7690 \pm 0.1672$ \\
DL & CNN & $0.8626 \pm 0.0974$ & $0.7252 \pm 0.1948$ & $0.8388 \pm 0.1262$ & 
$0.8254 
\pm 0.1395$ & $0.7395 \pm 0.1894$ \\
AL & SVMDC & $0.8635 \pm 0.0808$ & $0.7271 \pm 0.1616$ & $0.8334 \pm 0.1181$ & 
$0.8198 \pm 0.1244$ & $0.7454 \pm 0.1558$ \\
AL & CSSVM & $0.8689 \pm 0.0795$ & $0.7378 \pm 0.1590$ & $0.8435 \pm 0.1141$ & 
$0.8307 \pm 0.1207$ & $0.7548 \pm 0.1536$ \\
AL & WK-SMOTE & $0.7651 \pm 0.1131$ & $0.5302 \pm 0.2263$ & $0.6523 \pm 0.2231$ & 
$0.6250 \pm 0.2274$ & $0.5646 \pm 0.2269$ \\
AL & uNBSVM & $0.8315 \pm 0.0856$ & $0.6629 \pm 0.1712$ & $0.7927 \pm 0.1216$ & 
$0.7884 \pm 0.1253$ & $0.6852 \pm 0.1664$ \\
AL & RBI-LP-SVM & $0.5427 \pm 0.1139$ & $0.0855 \pm 0.2278$ & $0.0923 \pm 0.2294$ 
& 
$0.6926 \pm 0.0752$ & $0.0870 \pm 0.2279$ \\
AL & EBCS-SVM & $0.8784 \pm 0.1051$ & $0.7569 \pm 0.2102$ & $0.8581 \pm 0.1440$ & 
$0.8522 \pm 0.1502$ & $0.7671 \pm 0.2052$ \\
 \hline
\end{tabular}
\end{table*}

\begin{itemize}
\item Most methods reported results above 0.800 for BAR, GM, and uF1 scores, except 
for SVM for GM and uF1; WK-SMOTE for BAR, GM, and uF1; uNBSVM for GM and uF1; and 
RBI-LP-SVM for BAR, GM, and uF1.
\item Regarding BMI and uMCC, most methods reported performances above 0.700. The 
exceptions were SVM for BMI and WK-SMOTE, uNBSVM, and RBI-LP-SVM for BMI and uMCC.
\item ROS achieved the highest performance for BAR, BMI, GM, and uMCC and the 
second-best performance for uF1. RUS achieved the best performance for uF1 and the 
second-best performance for BAR, BMI, GM, and MCC. EBCS-SVM ranked the third-best 
for all metrics.
\item The hierarchical Bayesian analysis revealed a probability above 0.900 in the 
region of practical equivalence when EBCS-SVM was compared with ROS, SMOTE, 
SVMSMOTE, and RUS for all metrics. Thus, there is evidence in favor of the 
competitiveness of these methods for handling medium IR problems. We can observe 
this behavior in Fig.~\ref{fig:medium_posterior}, where we can note that the center 
mass is in the region of practical equivalence. Although for CNN and SVMDC, the 
center mass fell in the practically equivalent region, the posterior plot 
distribution was spread throughout the area of EBCS-SVM, providing at some extent 
evidence in favor of EBCS-SVM.
\item ROS stood out as the most effective data-level method. 
\item Among algorithm-level methods, EBCS-SVM obtained the best performance, and 
CSSVM was the second-best. Conversely, RBI-LP-SVM ranked in the last position.
\end{itemize}

For datasets with a medium IR, EBCS-SVM, ROS, and RUS were the most superior 
methods. ROS is highlighted as a prominent method for problems with medium IR. 
CSSVM remained a competitive method. In the next section, methods are evaluated 
when handling datasets with a high IR.

\subsection{Experiments with High Imbalance Ratio} \label{sec:high_ir}

\begin{figure*}[!ht]
\centering
\subfloat[SVM -- EBC-SVM]{
\includegraphics[width=0.23\textwidth]{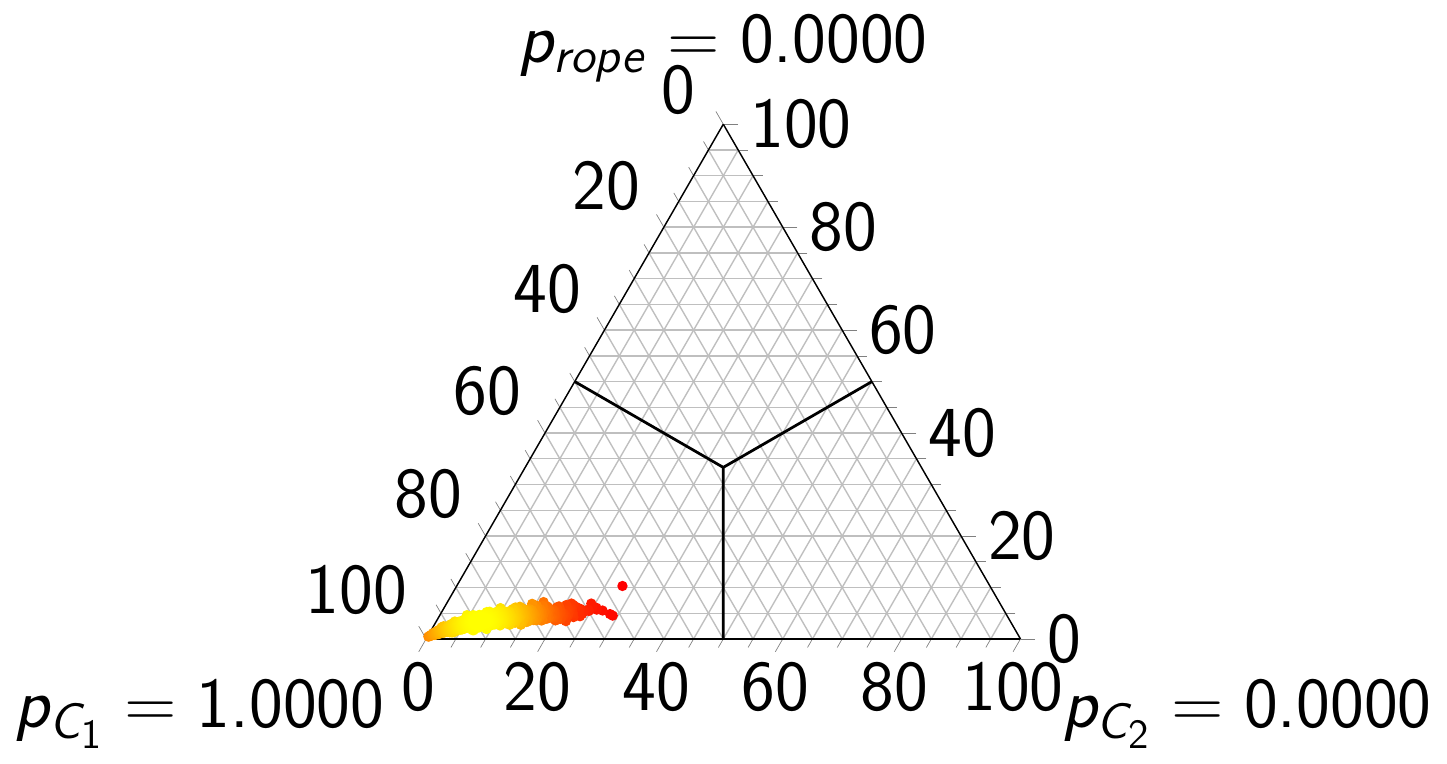}
\label{fig:high_svm_bar}}
\subfloat[ROS -- EBC-SVM]{
\includegraphics[width=0.23\textwidth]{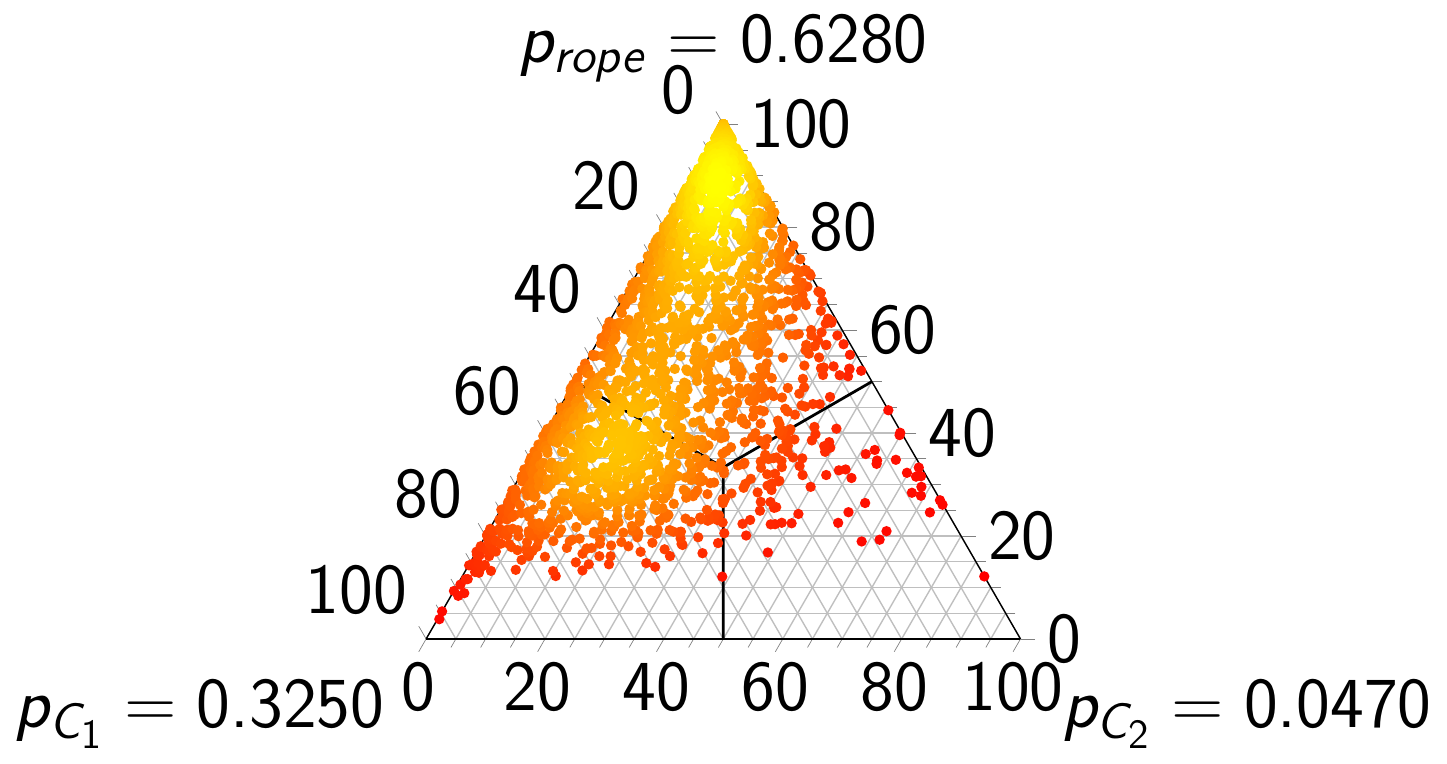}
\label{fig:high_ros_bar}}
\subfloat[SMOTE -- EBC-SVM]{
\includegraphics[width=0.23\textwidth]{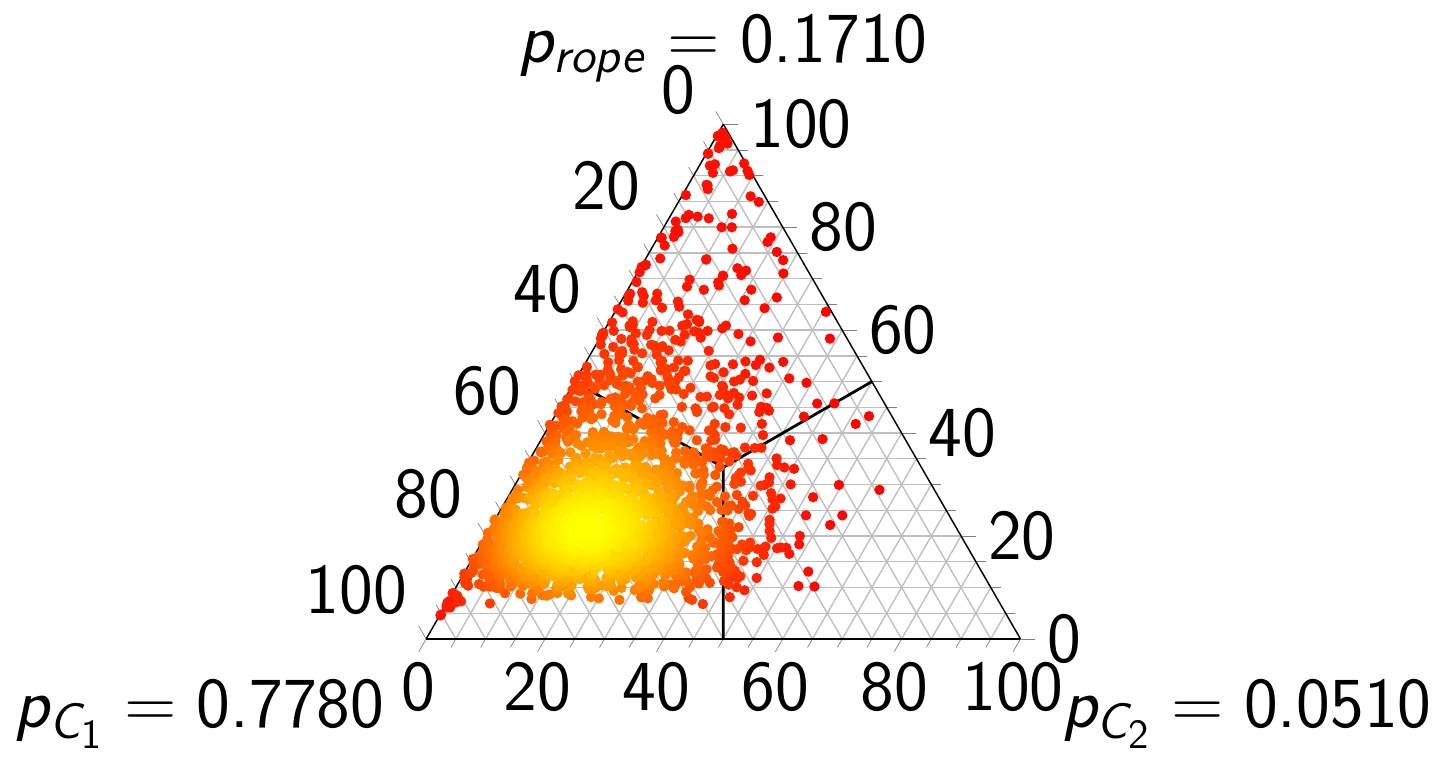}
\label{fig:high_smote_bar}}
\subfloat[SVMSMOTE -- EBC-SVM]{
\includegraphics[width=0.23\textwidth]{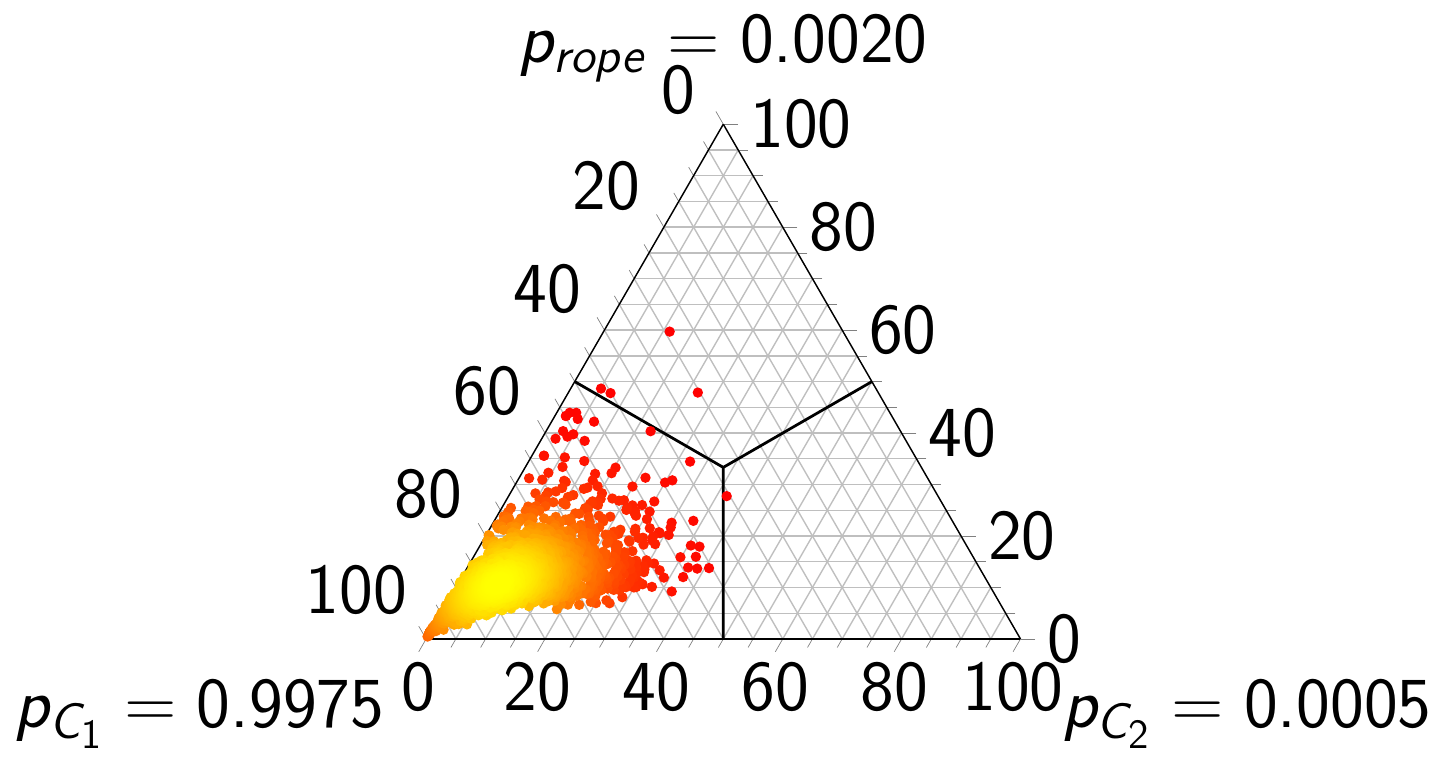}
\label{fig:high_svmsmote_bar}}
\qquad
\subfloat[RUS -- EBC-SVM]{
\includegraphics[width=0.23\textwidth]{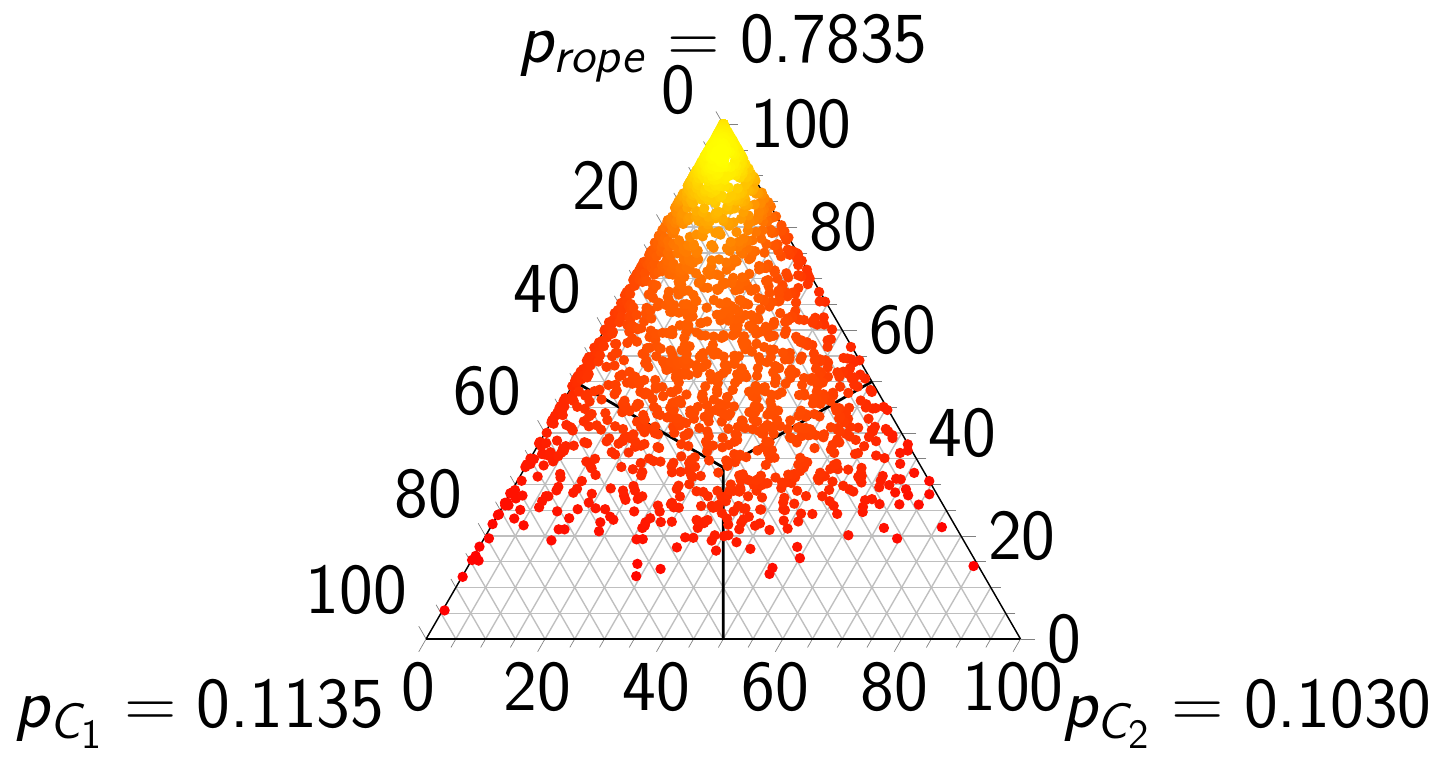}
\label{fig:high_rus_bar}}
\subfloat[CNN -- EBC-SVM]{
\includegraphics[width=0.23\textwidth]{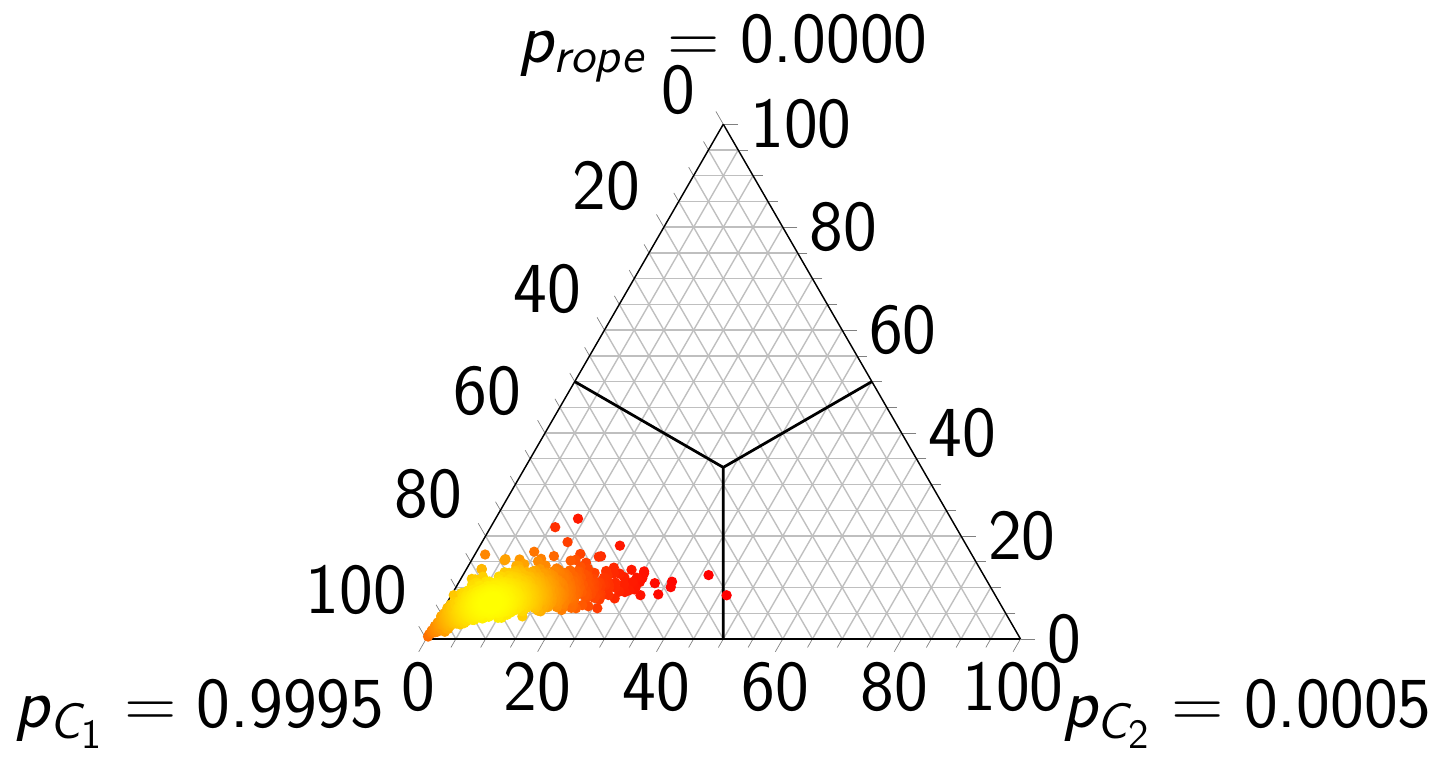}
\label{fig:high_cnn_bar}}
\subfloat[SVMDC -- EBC-SVM]{
\includegraphics[width=0.23\textwidth]{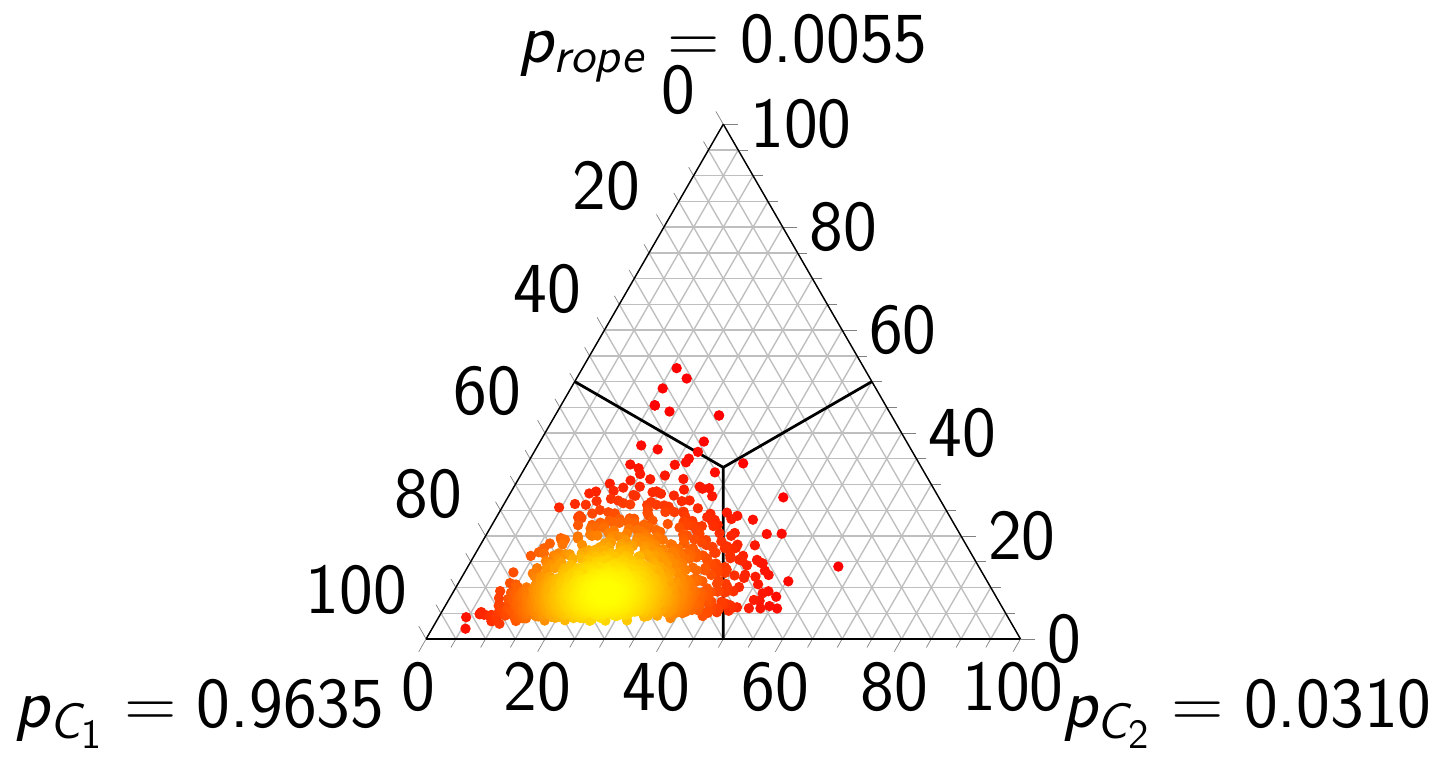}
\label{fig:high_svmdc_bar}}
\subfloat[CSSVM -- EBC-SVM]{
\includegraphics[width=0.23\textwidth]{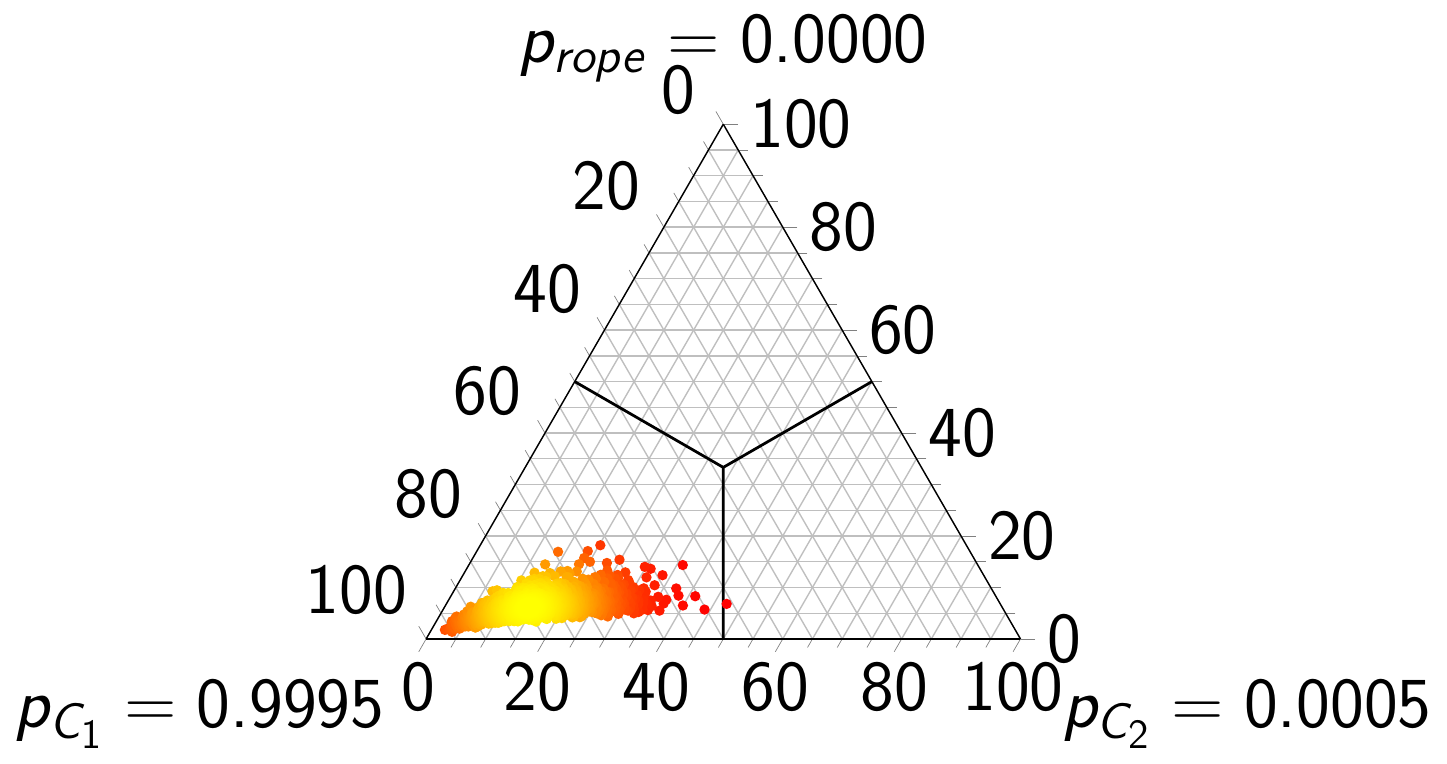}
\label{fig:high_cssvm_bar}}
\qquad
\subfloat[uNBSVM]{
\includegraphics[width=0.23\textwidth]{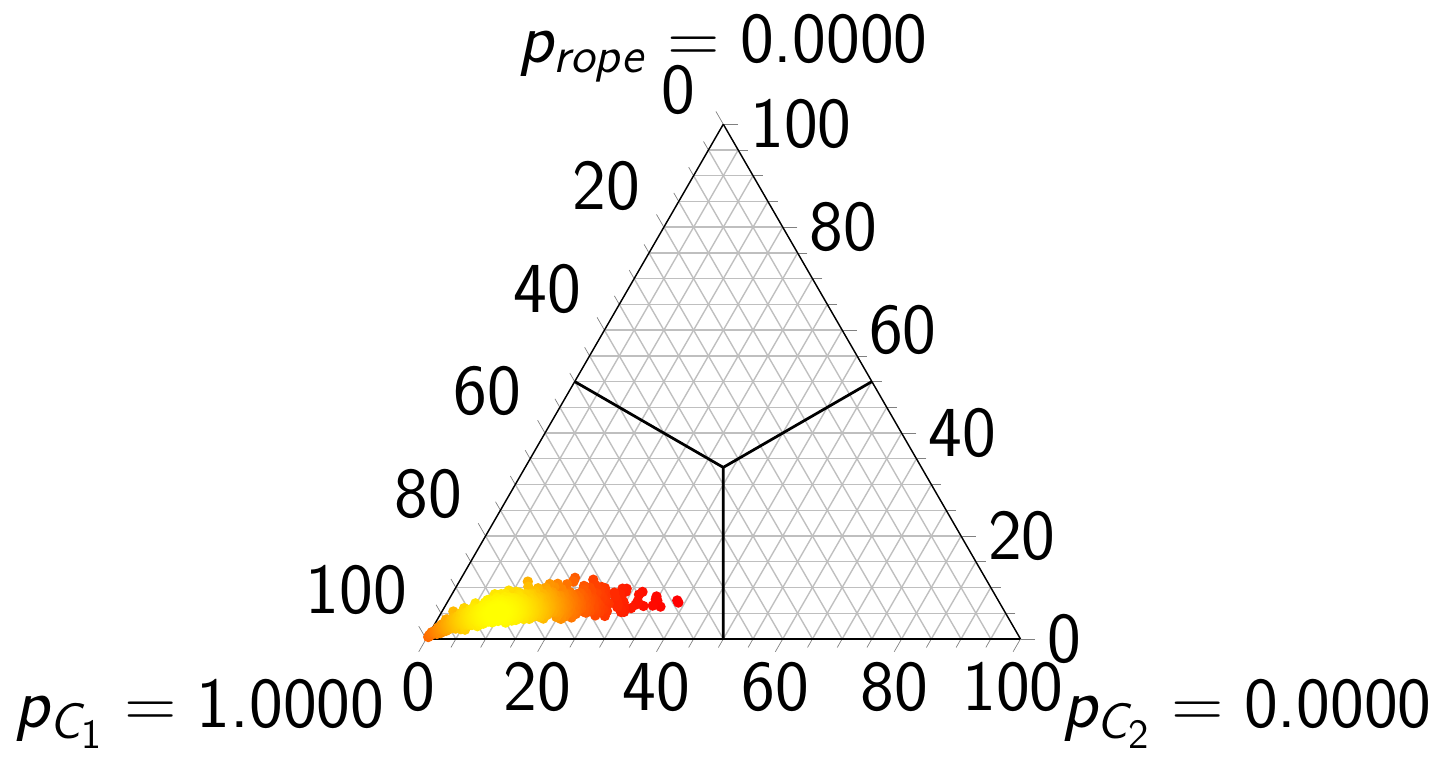}
\label{fig:high_unbsvm_bar}}
\subfloat[WK-SMOTE -- EBC-SVM]{
\includegraphics[width=0.23\textwidth]{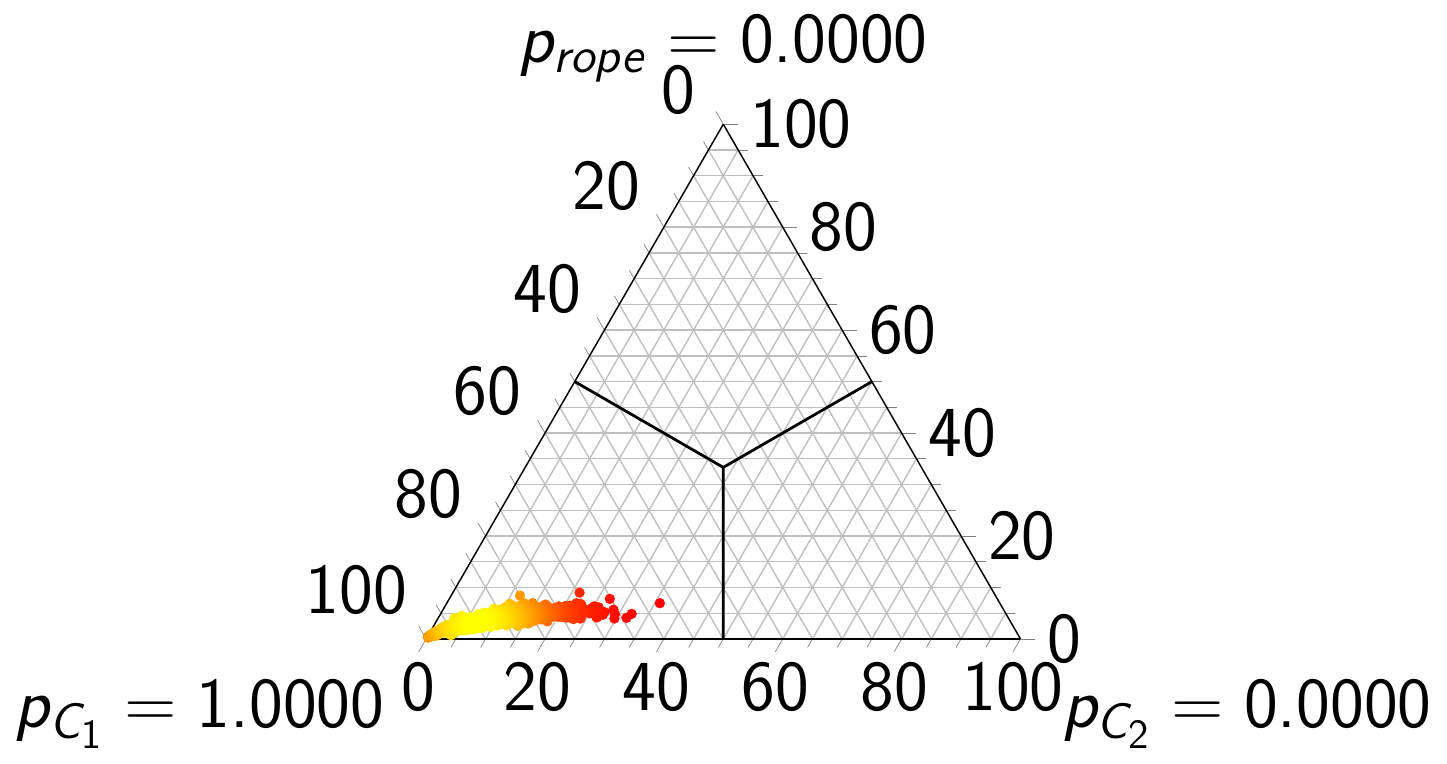}
\label{fig:high_wksmote_bar}}
\subfloat[RBI-LP-SVM -- EBC-SVM]{
\includegraphics[width=0.23\textwidth]{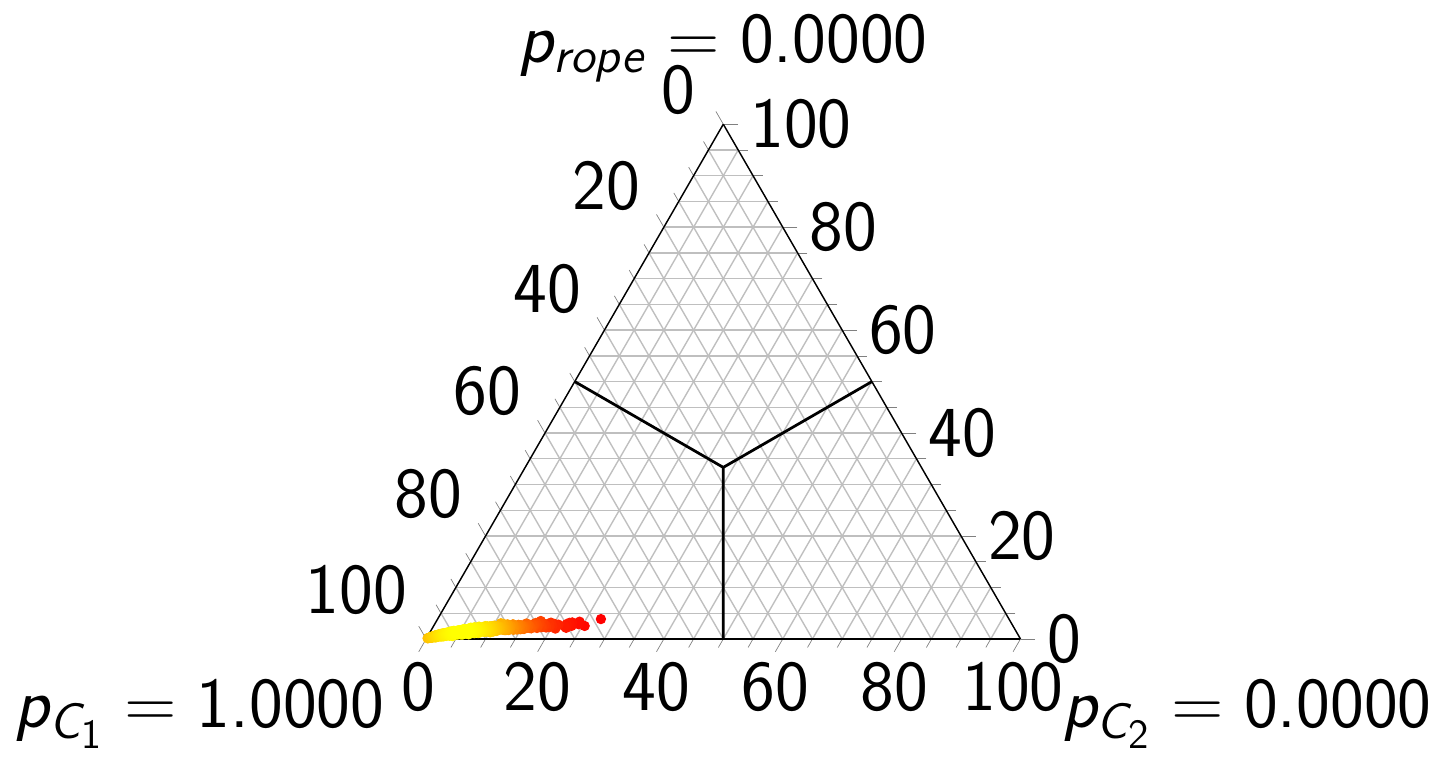}
\label{fig:high_rbi_bar}}
\caption{Posterior distribution for BAR when comparing EBCS-SVM with 
the reference methods on high IR datasetss. The region of the bottom-left 
represents EBCS-SVM, the region at the top is for rope, and the region in the 
bottom-right represents the reference method.}
\label{fig:high_posterior}
\end{figure*}

In this section, we considered the 25 datasets with an IR above 20 to analyze 
the performance of EBCS-SVM and reference methods. 
Table~\ref{tab:results_high_ir} reports the average results for all metrics, and 
Fig.~\ref{fig:high_posterior} shows the posterior probabilities obtained 
with the hierarchical Bayesian test. Based on these results, we remark the 
following:

\begin{table*}
\centering
 \caption{Obtained results on high IR datasets for BAR, BMI, GM, uF1, and 
uMCC scores.}
\label{tab:results_high_ir}
\begin{tabular}{llrrrrr}
\hline
 Fam. & Method & BAR & BMI & GM & uF1 & uMCC\\
 \hline
BL & SVM & $0.6712 \pm 0.1557$ & $0.3423 \pm 0.3114$ & $0.4413 \pm 0.3099$ & 
$0.4087 
\pm 0.3149$ & $0.3725 \pm 0.3105$ \\
DL & ROS & $0.7754 \pm 0.1119$ & $0.5508 \pm 0.2239$ & $0.6984 \pm 0.1770$ & 
$0.6762 
\pm 0.1844$ & $0.5665 \pm 0.2244$ \\
DL & SMOTE & $0.7644 \pm 0.1315$ & $0.5288 \pm 0.2629$ & $0.6665 \pm 0.2387$ & 
$0.6452 \pm 0.2426$ & $0.5436 \pm 0.2649$ \\
DL & SVMSMOTE & $0.7331 \pm 0.1437$ & $0.4661 \pm 0.2873$ & $0.5907 \pm 0.2766$ & 
$0.5652 \pm 0.2811$ & $0.4856 \pm 0.2880$ \\
DL & RUS & $0.7803 \pm 0.1132$ & $0.5606 \pm 0.2264$ & $0.7220 \pm 0.1726$ & 
$0.7113 
\pm 0.1716$ & $0.5692 \pm 0.2274$ \\
DL & CNN & $0.7203 \pm 0.1412$ & $0.4406 \pm 0.2824$ & $0.6030 \pm 0.2297$ & 
$0.5740 
\pm 0.2391$ & $0.4535 \pm 0.2899$ \\
AL & SVMDC & $0.7340 \pm 0.1500$ & $0.4680 \pm 0.3000$ & $0.5825 \pm 0.3144$ & 
$0.5655 \pm 0.3118$ & $0.4849 \pm 0.3037$ \\
AL & CSSVM & $0.7031 \pm 0.1640$ & $0.4062 \pm 0.3280$ & $0.5003 \pm 0.3371$ & 
$0.4773 \pm 0.3394$ & $0.4257 \pm 0.3309$ \\
AL & WK-SMOTE & $0.6604 \pm 0.1415$ & $0.3208 \pm 0.2830$ & $0.4119 \pm 0.3016$ & 
$0.3964 \pm 0.3024$ & $0.3392 \pm 0.2877$ \\
AL & uNBSVM & $0.7054 \pm 0.1321$ & $0.4108 \pm 0.2642$ & $0.5492 \pm 0.2659$ & 
$0.5410 \pm 0.2642$ & $0.4278 \pm 0.2678$ \\
AL & RBI-LP-SVM & $0.5470 \pm 0.1034$ & $0.0940 \pm 0.2069$ & $0.1440 \pm 0.2709$ 
& 
$0.6410 \pm 0.1298$ & $0.0977 \pm 0.2153$ \\
AL & EBCS-SVM & $\mathbf{0.7972 \pm 0.1234}$ & $\mathbf{0.5944 \pm 0.2468}$ & 
$\mathbf{0.7482 \pm 0.1732}$ & $\mathbf{0.7430 \pm 0.1757}$ & $\mathbf{0.6079 
\pm 0.2431}$ \\
 \hline
\end{tabular}
\end{table*}

\begin{itemize}
\item EBCS-SVM excelled in dealing effectively with highly imbalanced datasets in 
all metrics. Ergo, the superiority of EBCS-SVM is stressed when the IR is 
increased. 
\item Among data-level methods, RUS had the best score in all metrics and ranked 
second-best position globally. ROS was the second-best 
data-level method and ranked third-best position globally.
\item Among algorithm-level methods, EBCS-SVM was the best one, followed by SVMDC. 
However, when observing the posterior probabilities reported by the 
hierarchical Bayesian test, we noted that the center mass is in the region of 
EBCS-SVM. Furthermore, the posterior odds revealed strong evidence in favor of 
EBCS-SVM. We observed this behavior for all metrics. 
\item Posterior probabilities plots also showed that, for most reference 
methods, the center mass fell in the region of EBCS-SVM. The exception was RUS, 
whose center mass is in the region of practical equivalence; however, its 
distribution spreads in the regions of both EBCS-SVM and RUS.
\end{itemize}

As the IR increased, the performance of EBCS-SVM stood out over the reference 
methods, regardless of the adopted metric. The hierarchical Bayesian test 
reported high probabilities in favor of EBCS-SVM that supported these 
observations.

\subsection{Computational Time} \label{sec:computational_time}

\begin{figure*}
\centering
\resizebox{0.89\textwidth}{!}{
\includegraphics{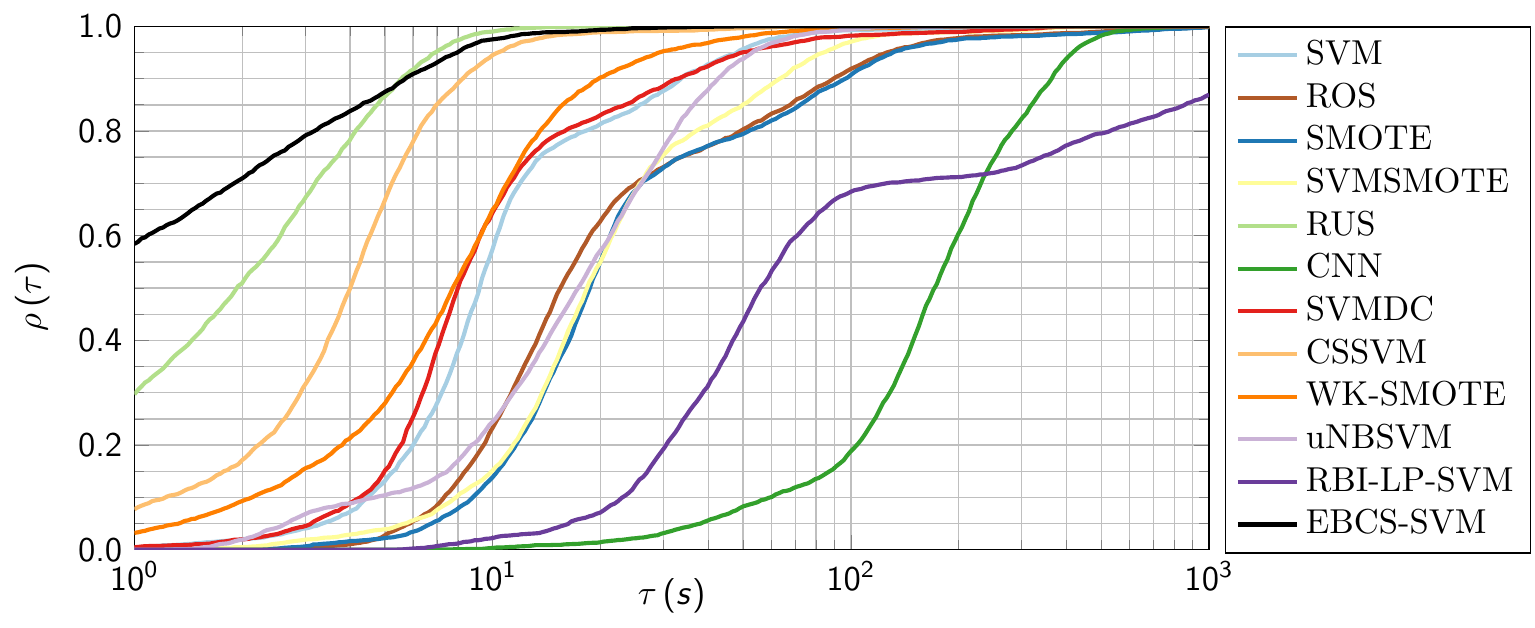}}
\caption{Performance profile of the training time for each method.}
\label{fig:performance_profile}
\end{figure*}

In this section, we analyze the training time for each method. We used the 
performance profile~\cite{Dolan2002}, which represents the cumulative 
distribution on a performance metric. The performance profile is constructed for 
the necessary training time required by each method to optimize the 
hyper-parameters and learn the classification model. 
Fig.~\ref{fig:performance_profile} depicts the performance profile for all 
methods. The $y-$axis represents the probability 
$\left(\rho\left(\tau\right)\right)$ that a method can learn a model within a 
factor $\tau$ times the fastest method, and the $x-$axis represents the $\tau$ 
factor. Thus, $\rho\left(1\right)$ indicates the probability where a given method 
achieves the lowest training time among all methods.

From Fig.~\ref{fig:performance_profile}, we can observe that both RUS and 
EBCS-SVM exhibited the best training times. From the value of $\rho\left(1\right)$, 
it is observed that EBCS-SVM had the highest probability $\left(0.58\right)$ of 
being the fastest one, while RUS had the second-highest probability 
$\left(0.30\right)$. Furthermore, algorithm-level methods generally required 
less training time than data-level methods, which can be observed in the near-zero 
probability of oversampling techniques with a value of $\tau$ equals one. The 
outstanding performance of RUS in training time is because by removing training 
instances, the SVM algorithm works with a reduced number of samples and reduces the 
computational time. Although CNN is also an undersampling method, the way in how 
instances to remove are selected slows down the training time. On the other hand, 
EBCS-SVM exploits the information of both levels to feedback with the information 
of previous support vectors of similar solutions and improving the convergence in 
solving the lower-level. Finally, RBI-LP-SVM showed the worst performance profile.

\section{Conclusions} \label{sec:conclusions}

This paper introduced EBCS-SVM for learning an SVM in imbalanced scenarios. 
EBCS-SVM formulated the optimization of hyperparameters and support vectors as a 
bilevel optimization problem and showed to be able to handle imbalanced 
classification problems effectively and freed practitioners from defining the 
optimal cost to each class. To this end, an EA at the upper-level and the SMO at 
the lower-level interplay, such that lower-level solutions impact the BER of the 
upper-level solutions, and previous hyper-parameters help initialize the set of 
support vectors. Thus, there is a dual enrichment in the two levels.

The efficacy of EBCS-SVM was assessed using 70 benchmark datasets and compared with 
those of state-of-the-art techniques. Experimental results revealed a leading 
performance of EBCS-SVM. The traditional SVM was unable to deal effectively with 
imbalanced datasets. On the other hand, data-level methods showed excellent 
performance, although in most cases required larger training times than 
algorithm-level methods. SVMSMOTE and CNN were the worst among data-level methods. 
The low performance of both can be because the implicit mapping of the kernel 
function is not taken into account to select boundary instances when sampling. 
EBCS-SVM considered the particularities of SVM to learn a model.

The most competitive method was RUS, a data-level technique that randomly 
subsamples the majority class. Since it does not require further information, RUS 
is fast. However, this method was outperformed by EBCS-SVM as the imbalanced ratio 
increased and EBCS-SVM required lower training time. Moreover, EBCS-SVM employed a 
self-adapted EA to adjust the evolutionary parameters during the learning. 
Therefore, EBCS-SVM is accurate and does not require fine-tuning of the SVM 
hyperparameters.

\bibliographystyle{IEEEtran}
\bibliography{references}

\begin{IEEEbiography}[{\includegraphics[width=1in,height=1.25in,clip,keepaspectratio]{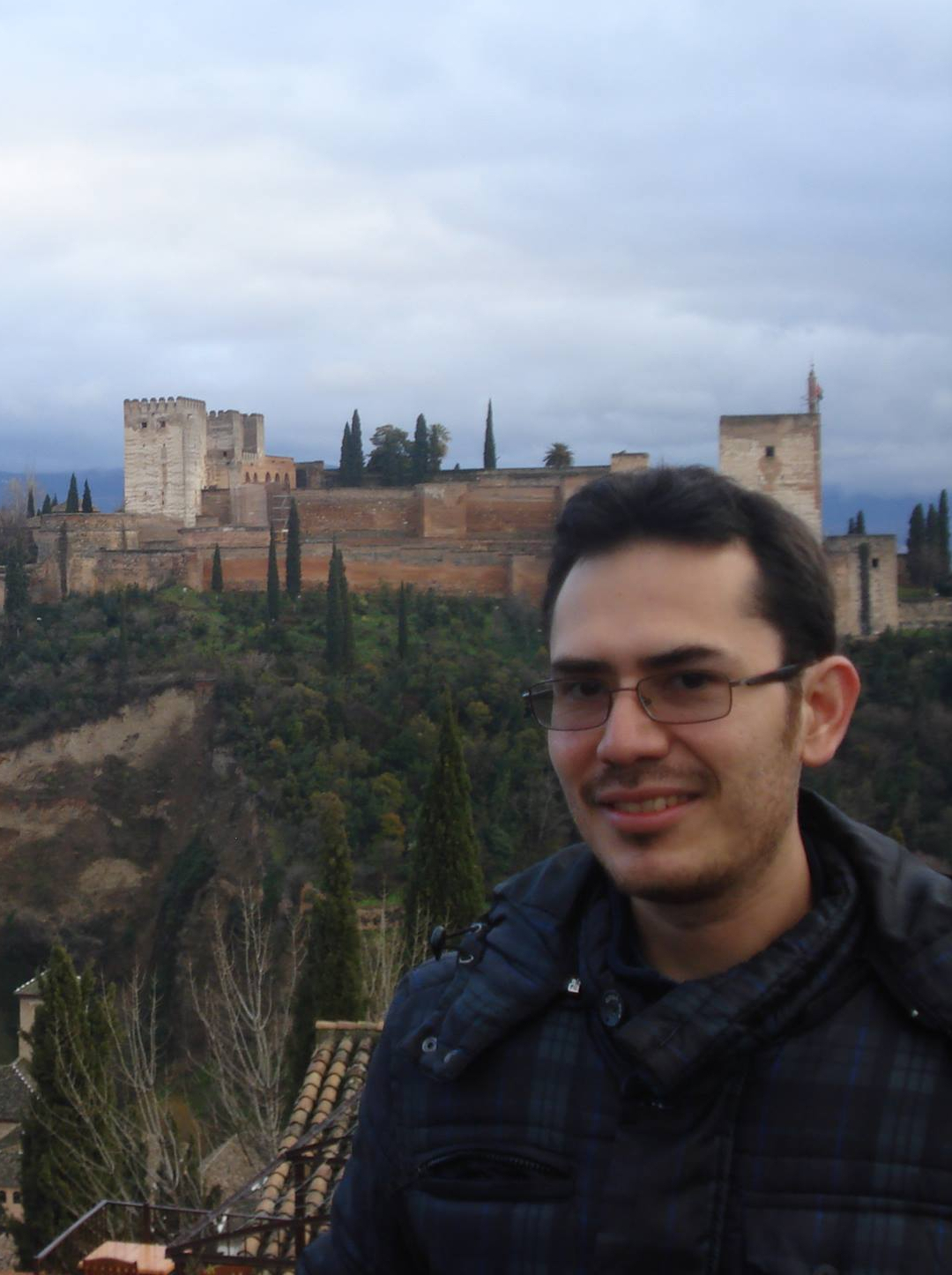}}]{Alejandro Rosales-P{\'e}rez}
received the B.S. degree in electronic engineering from the Instituto 
Tecnol{\'o}gico de Tuxtla Guti{\'e}rrez, Chiapas, Mexico in 2008, and the M.Sc. and 
Ph.D. degrees in computer science from INAOE, located at Puebla, Mexico in 2011 
and 2016, 
respectively. He is currently with CIMAT Monterrey. His doctoral thesis was 
awarded by the National Association of Education Institutions in Information 
Technology (ANIEI) in 2016 and also by the Mexican Society on Artificial 
Intelligence (SMIA) in 2016. He is member of the Mexican National System of 
Researchers (SNI). His current research interests mainly include evolutionary 
computation and machine learning.
\end{IEEEbiography}

\begin{IEEEbiography}[{\includegraphics[width=1in,height=1.25in,clip,
keepaspectratio]{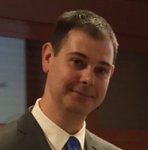}}]{Salvador Garc{\'i}a}
received the B.S. and Ph.D. degrees in Computer Science from the University of 
Granada, Granada, Spain, in 2004 and 2008, respectively. He is currently a Full 
Professor in the Department of Computer Science and Artificial Intelligence, 
University of Granada, Granada, Spain. Dr. García has published more than 100 
papers in international journals (more than 70 in Q1), h-index 54. He is an 
associate editor in chief of “Information Fusion” (Elsevier), and an associate 
editor of “Swarm and Evolutionary Computation” (Elsevier) and “AI Communications” 
(IOS Press) journals. His research interests include data science, data 
preprocessing, Big Data, evolutionary learning, Deep Learning and metaheuristics. 
He belongs to the list of the Highly Cited Researchers in the area of Computer 
Sciences (2014-2020): \url{http://highlycited.com/} (Clarivate Analytics). 
\end{IEEEbiography}

\begin{IEEEbiography}[{\includegraphics[width=1in,height=1.25in,clip,keepaspectratio]{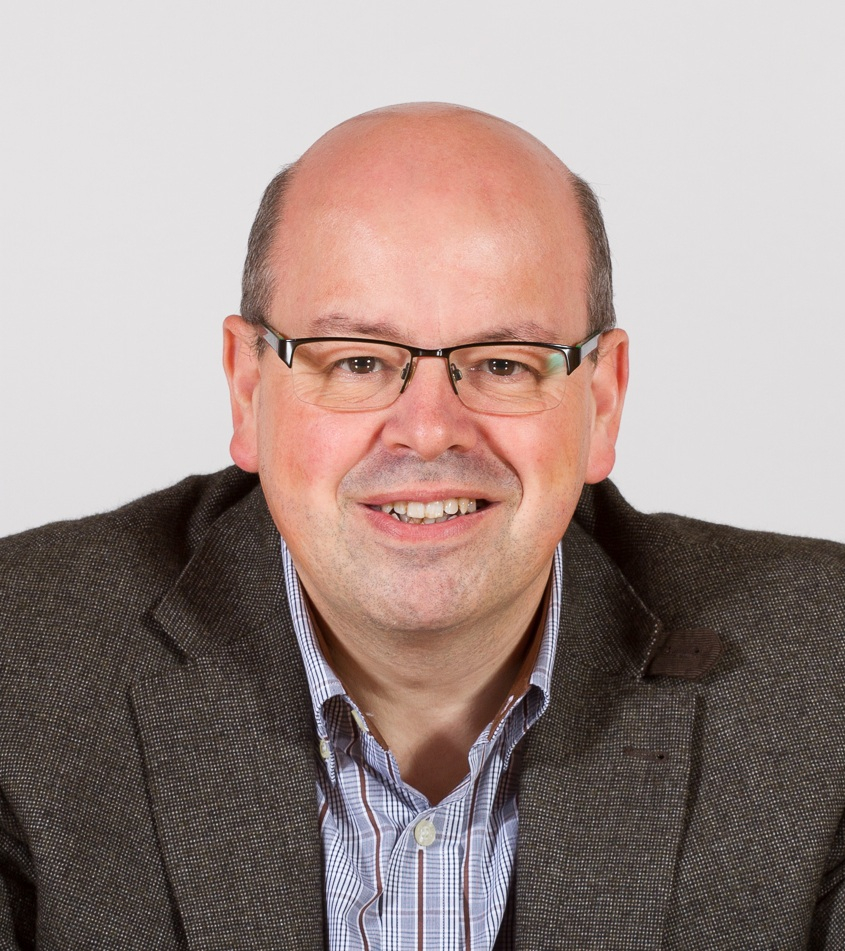}}]{Francisco Herrera}
(SM'15) received  his  M.Sc.  in Mathematics  in  1988  and  Ph.D.  in  
Mathematics  in 1991,  both  from  the  University  of  Granada,  Spain. He  is  
currently  a  Professor  in  the  Department  of Computer  Science  and  Artificial 
 Intelligence  at  the University  of  Granada  and  Director  of  DaSCI  Institute 
 (Andalusian  Research  Institute  in  Data  Science  and  Computational  
Intelligence).  His  current research  interests  include  among  others,  
Computational Intelligence (including fuzzy modeling, computing with words, 
evolutionary algorithms and deep learning), information fusion and decision making, 
and data science (including data  preprocessing,  prediction  and  big  data).  He  
has  been  the  supervisor  of more than 50 Ph.D. students. He has published more 
than 500 journal papers, receiving  more  than  98,000  citations  (Google  
Scholar,  H-index  152).  He  is co-author  of  several  books,  and  the  Editor  
in  Chief  of  Information  Fusion (Elsevier). He is been selected as a Highly 
Cited Researcher (in the fields of Computer Science and Engineering, 2014 to 
present, Clarivate Analytics)
\end{IEEEbiography}

\end{document}